\documentclass[runningheads]{llncs}
\usepackage{graphicx}

\usepackage{tikz}
\usepackage{comment}
\usepackage{amsmath,amssymb} 
\usepackage{color}
\usepackage{booktabs}
\usepackage{multirow}
\usepackage{colortbl}
\usepackage{bm}
\usepackage{subcaption}
\usepackage{xcolor}
\usepackage{varwidth}

\usepackage[accsupp]{axessibility}  

\usepackage{cite}
\usepackage[pagebackref=true,breaklinks=true,letterpaper=true,colorlinks,bookmarks=false]{hyperref}

\usepackage{xspace}
\usepackage{orcidlink}

\makeatletter
\DeclareRobustCommand\onedot{\futurelet\@let@token\@onedot}
\def\@onedot{\ifx\@let@token.\else.\null\fi\xspace}

\def\eg{\emph{e.g}\onedot} 
\def\ie{\emph{i.e}\onedot}

\makeatother

\begin{document}
\pagestyle{headings}
\mainmatter
\def\ECCVSubNumber{6145}  

\title{Deep Semantic Statistics Matching (D2SM) Denoising Network} 

\author{Kangfu Mei\orcidlink{0000-0001-8949-9597}\ \and
Vishal M. Patel \and
Rui Huang}

\authorrunning{Mei et al.}

\institute{Johns Hopkins University \and The Chinese University of Hong Kong, Shenzhen\\
\url{https://kfmei.page/d2sm}}

\maketitle

\begin{abstract}
  The ultimate aim of image restoration like denoising is to find an exact correlation between the noisy and clear image domains.
  But the optimization of end-to-end denoising learning like pixel-wise losses is performed in a sample-to-sample manner, which ignores the intrinsic correlation of images, especially semantics.
  In this paper, we introduce the Deep Semantic Statistics Matching (D2SM) Denoising Network.
  It exploits semantic features of pretrained classification networks, then it implicitly matches the probabilistic distribution of clear images at the semantic feature space.
  By learning to preserve the semantic distribution of denoised images, we empirically find our method significantly improves the denoising capabilities of networks, and the denoised results can be better understood by high-level vision tasks.
  Comprehensive experiments conducted on the noisy Cityscapes dataset demonstrate the superiority of our method on both the denoising performance and semantic segmentation accuracy.
  Moreover, the performance improvement observed on our extended tasks including super-resolution and dehazing experiments shows its potentiality as a new general plug-and-play component.
\end{abstract}

\section{Introduction}
Deep learning based methods~\cite{dong2015image, zhang_ffdnet_2018, li2018multi} have achieved a dramatic leap in the performance of various image restoration tasks.
Typically, they employ a Convolutional Neural Network (CNN) on a set of image pairs, consisting of degraded images and corresponding clear images, for restoration learning.
By maximizing the correspondence between each pair of the CNN-restored results and the clear image, the CNN is trained to map images from the degraded domain into the clear domain.
However, blur issues always existed in such a manner.
Recent work~\cite{johnson_perceptual_2016} called perceptual loss finds that maximizing the correspondence in the semantic feature space of pre-trained large-scale classification network (\emph{e.g.}, VGG~\cite{simonyan_very_2015} network trained on ImageNet~\cite{deng_imagenet_2009}) leads to better visual quality~\cite{zhang_unreasonable_2018, mechrez_maintaining_2018}.
A more widely used strategy inspired by GANs~\cite{goodfellow_generative_2014}, which employs a discriminator to implicitly enforce the distribution of restored images to be consistent with the distribution of clear images in terms of KL- and JS- divergence, can largely improve the perceptual quality of restored images.
But the training procedure is often unstable, mostly because the objective is a zero-sum non-cooperative game that cannot be easily solved.
Thus, it is straightforward to wonder whether it is possible to combine the pre-trained large-scale networks in an adversarial or statistical manner to bypass their drawbacks and avail their advantages together?

\begin{figure*}[t]
  \begin{subfigure}[c]{.22\linewidth}
    \captionsetup{justification=centering, labelformat=empty, font=small}
    \includegraphics[width=1\linewidth]{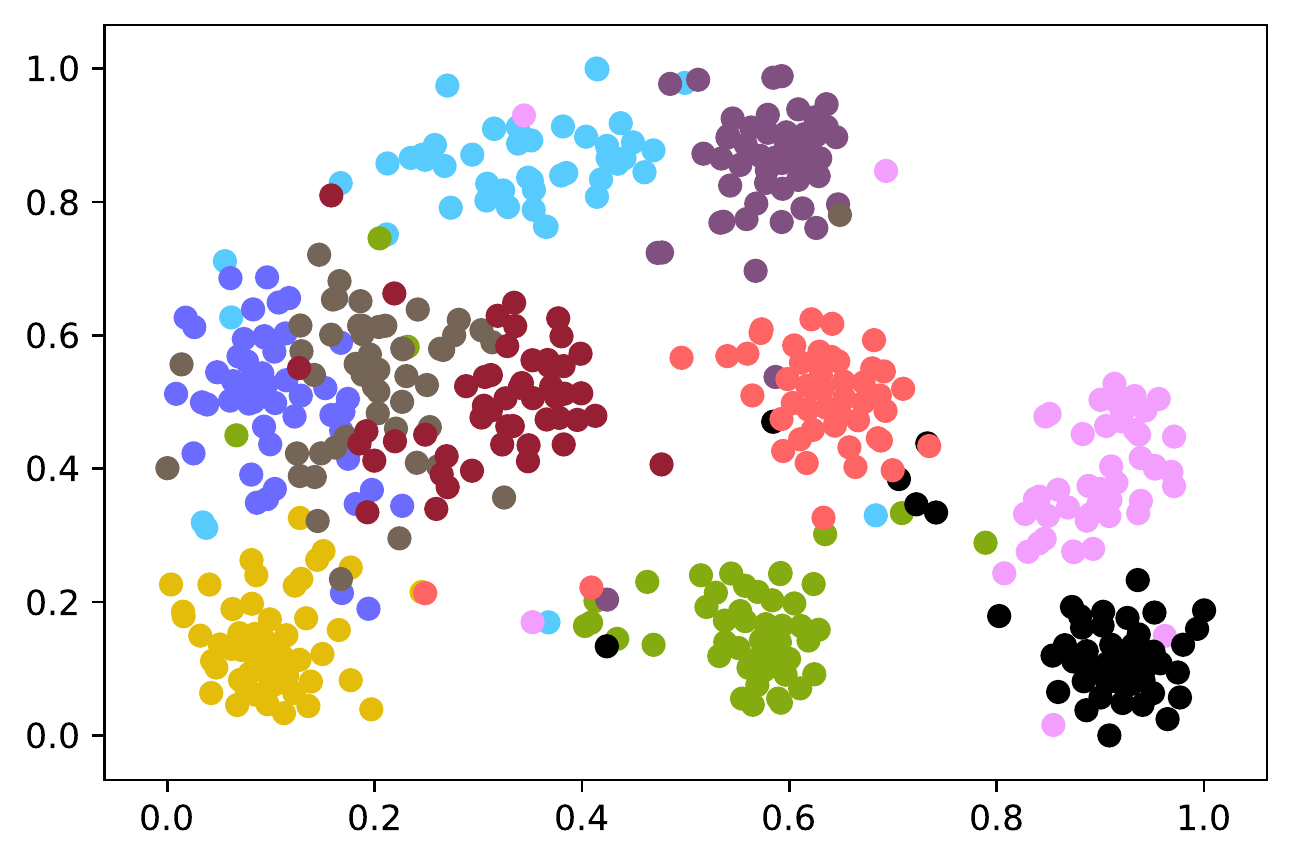}
    \caption{w. $\mathcal{L}_1$}
  \end{subfigure}
  \hfill
  \begin{subfigure}[c]{.22\linewidth}
    \captionsetup{justification=centering, labelformat=empty, font=small}
    \includegraphics[width=1\linewidth]{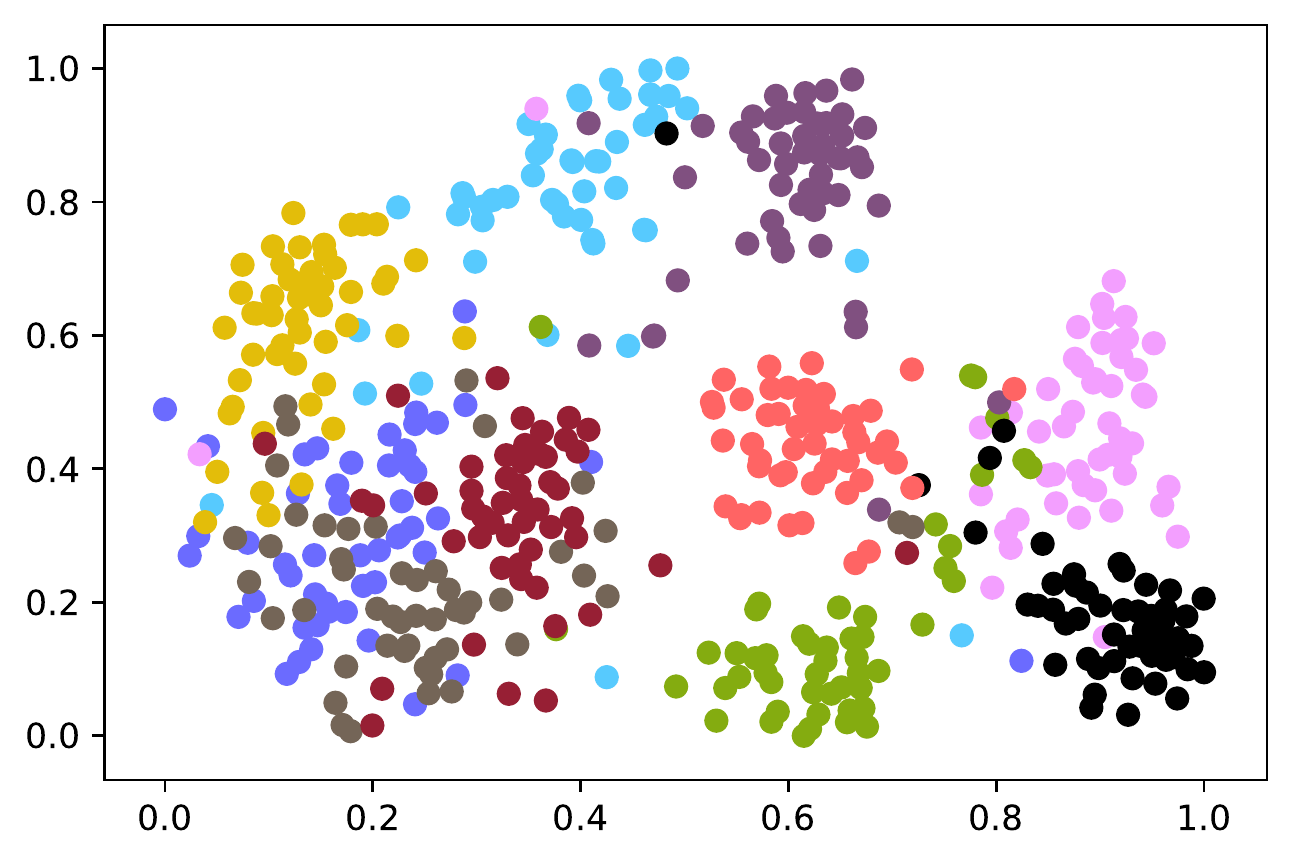}
    \caption{w. $\mathcal{L}_{Perceptual}$}
  \end{subfigure}
  \hfill
  \begin{subfigure}[c]{.22\linewidth}
    \captionsetup{justification=centering, labelformat=empty, font=small}
    \includegraphics[width=1\linewidth]{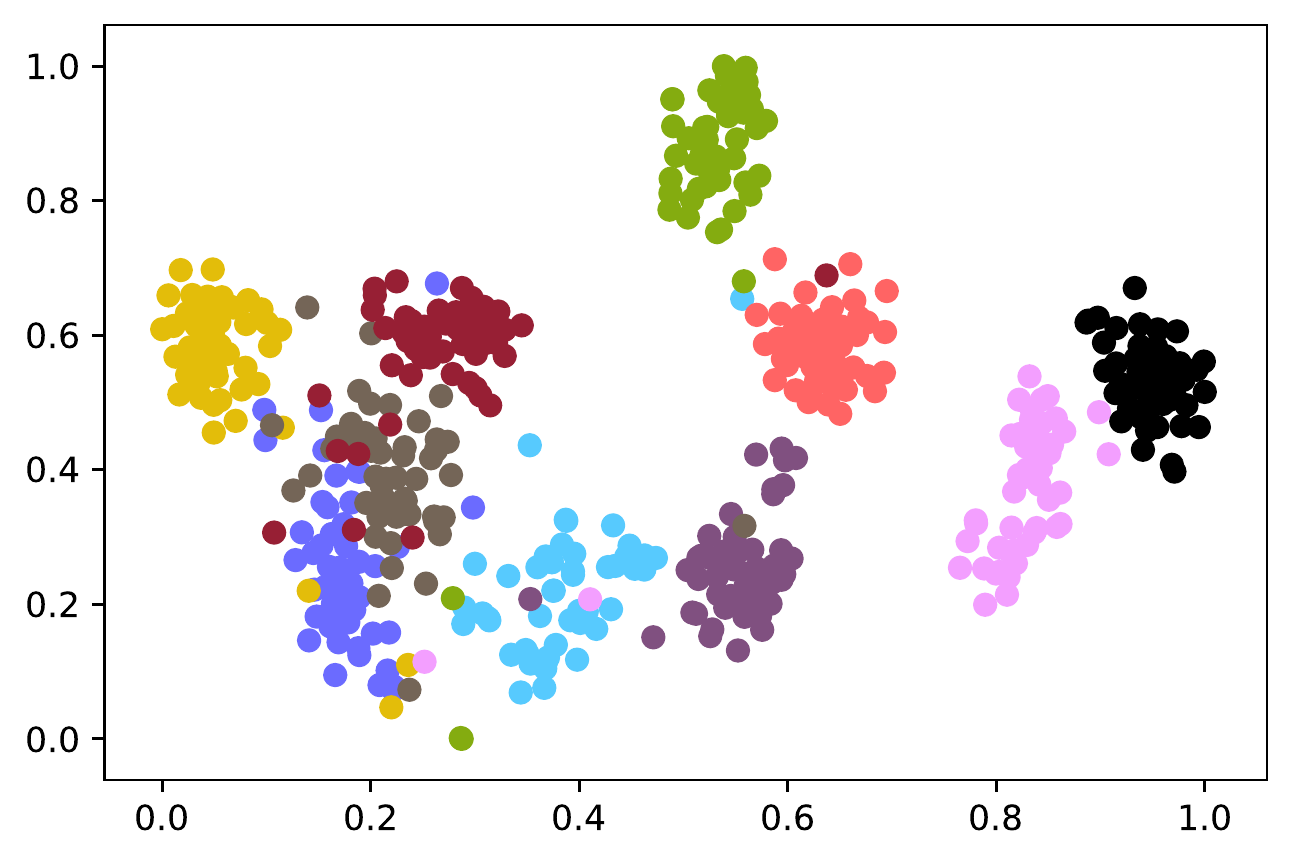}
    \caption{Ours}
  \end{subfigure}
  \hfill
  \begin{subfigure}[c]{.285\linewidth}
    \captionsetup{justification=centering, labelformat=empty, font=small}
    \includegraphics[width=1\linewidth]{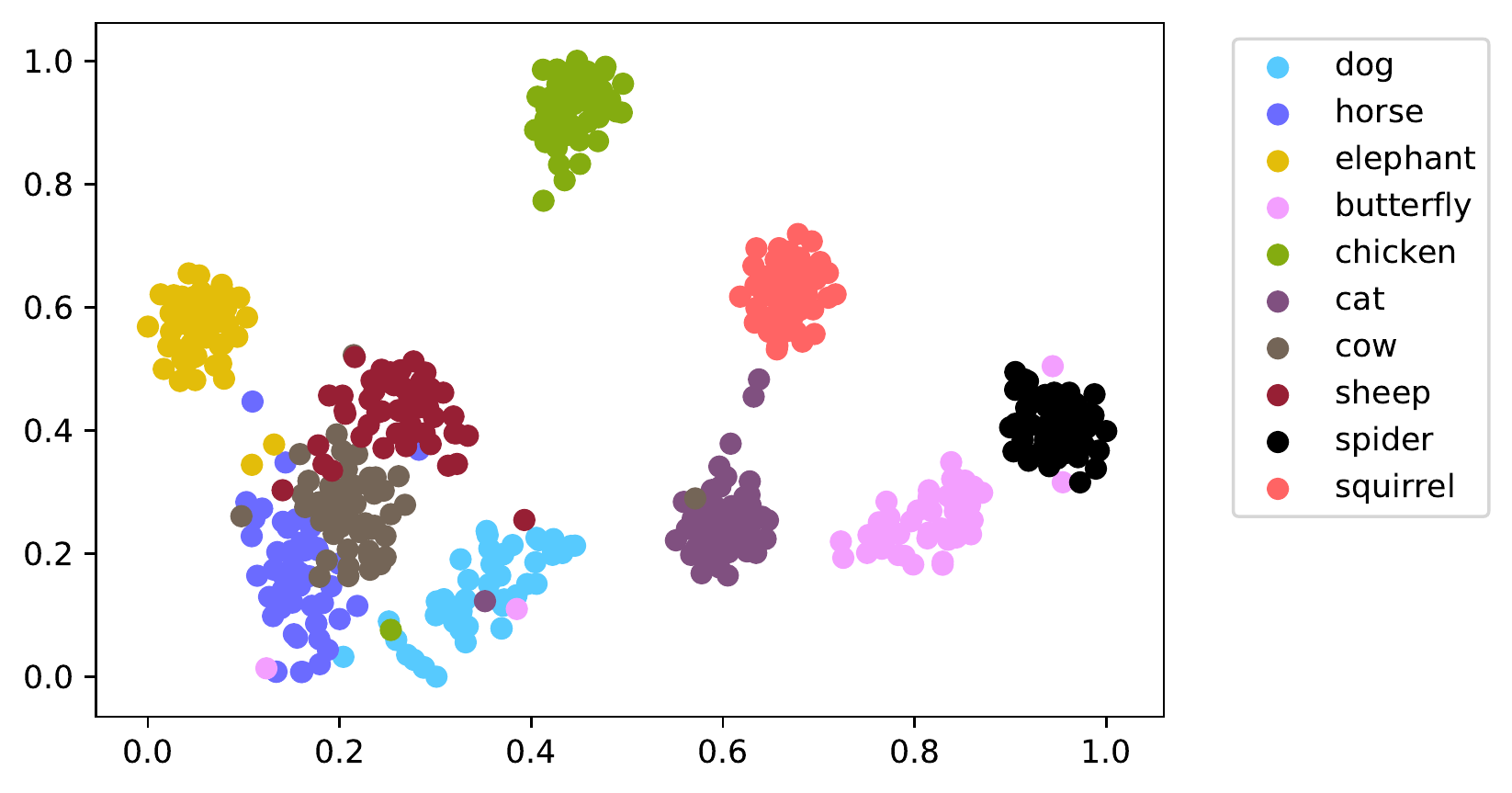}
    \caption{Clear Images}
  \end{subfigure}  
  \caption{\textbf{t-SNE of Denoised Images in the Semantic Feature Space}
    By exploiting t-SNE~\cite{van_der_maaten_visualizing_2008} to reduce dimensions of semantic features and project them into 2D coordinates, we visualize the distributions of denoised animal images in the semantic feature space. Ours preserves most semantics as the clear images.}
  \label{fig:distribution}
\end{figure*}

To answer the above question, we first look at the training procedure from a probabilistic view in the semantic feature space (the space of extracted semantic features of each images), which contains many clusters of different semantics.
Figure~\ref{fig:distribution} visualizes these clusters with t-SNE~\cite{van_der_maaten_visualizing_2008} in animals images selected from ImageNet~\cite{deng_imagenet_2009}.
An image that belongs to a specific cluster and owns intrinsic semantics is called Single-Semantic Image in this paper.
For example, animal images are single-semantic images, because they have common semantics of species, even though individual animals with the same specie look different.
Intuitively, restored single-semantic images should preserve the same distribution as the corresponding clear single-semantic images in the semantic feature space.
However, the objectives of existed methods for denoising learning cannot preserve this, as w.$\mathcal{L}_1$ and w.$\mathcal{L}_{Perceptual}$ show in Figure~\ref{fig:distribution}.
Therefore, we argue that minimizing the divergence between the probability distributions estimated in the restored domain and the one estimated in the clear domain should be a potential promising solution.
Similar idea is also validated by MMDGAN~\cite{li2017mmd} and its variants for generating face and bedroom images, but the idea is rarely researched in the restoration literature.

Different from the single-semantic image, natural images like a cityscape image often consists of multiple objects, and can not be easily identified by a single semantic, here it is called Complex-Semantic image.
The semantic feature extracted from such an image may not belong to any simple semantic clusters, but resides on an extremely complicated manifold in the semantic space.
For example, cityscape images usually consist of objects/regions of different semantics like \emph{road}, \emph{sidewalk}, and \emph{building}, as classified by the Cityscapes dataset~\cite{cordts_cityscapes_2016}.
To approximate and compare the probability distributions of complex-semantic images is nontrivial due to their unique intrinsic uncertainty of semantics.

In the paper, we propose a new distribution-wise objective for denoising learning, and is capable of being extended into general restoration tasks, towards both the single-semantic images and complex-semantic images. 
It learns to preserve the probability distribution of denoised images in the semantic feature space, and it is called as Deep Semantic Statistics Matching (D2SM) Denoising Network.
The objective of D2SM exploits a way similar to Kernel Density Estimation (KDE)~\cite{scott_multivariate_2015} to implicitly estimate the probability distributions of semantic features from a set of denoised images and clear images, and then Kullback-Leibler (KL) divergence between two distributions is used as the objective.
Here, one of our major novelty comes from the way of density estimation, where we model the probability distributions based on internal patches from a single complex-semantic image or multiple single-semantic images.
The way of availing internal patches tends to be more appropriate than modeling the distribution of multiple complex-semantic images.
Such a phenomenon is also proved in recent work~\cite{zontak_internal_2011} suggests that the internal visual entropy of a single image is much smaller than multiple images.
Therefore, we propose to use the divergence of patch distributions to guide the learning, called Patch-Wise Internal Probability.

Nevertheless, statistically estimating the density of patches conducted in a single mini-batch usually requires a great large number of samples.
To maintain the trade-off between the computational cost and accuracy, another major novelty of the paper, called Memorized Historic Sampling, is proposed, inspired by recent contrastive learning related works~\cite{wu_unsupervised_2018,he_momentum_2020}.
By simply leveraging the statistics among the mini-batch and memorized historic mini-batch in queues, we demonstrate that D2SM significantly outperforms the same network backbone with perceptual loss and other state-of-the-art objectives, without additional information or parameters.
Empirical evaluation validates that D2SM largely improves not only the effectiveness of denoising, but also super-resolution and dehazing, and hence it should be able to be generally applied to different tasks and network architectures.\\

\noindent \underline{Our contributions are therefore three-fold:} \\
(i) We propose D2SM for image denoising learning, which minimizes the distribution divergence instead of the sample-to-sample distance in the semantic feature space.
(ii) D2SM is adapted to complex-semantic images in a patch-wise manner, which can decompose complex semantics in natural images for efficient distribution approximation.
(iii) Extensive experiments are conducted to demonstrate that D2SM substantially outperforms the original perceptual loss and other state-of-the-art losses, without modifying the network architecture or accessing the additional data.
The superior accuracy in high-level vision tasks further validates that D2SM indeed transfers semantics for restoration.

\section{Related Work}
Resulted by the emergence of deep neural networks, recent CNN based methods have led to a dramatic leap in image restoration.
Among them, most works utilize the pixel-wise similarity metrics as their objective, \emph{e.g.}, $\mathcal{L}_1$ and $\mathcal{L}_{MSE}$.
Though higher performance in metrics like PSNR or SSIM~\cite{wang_image_2004} is achieved by using these loss functions, recent work~\cite{zhang_unreasonable_2018} finds that these metrics do not reflect human perceptual preferences.
In contrast, results generated by CNNs trained with the perceptual objective are more closely correlated with the human judgment~\cite{mechrez_maintaining_2018}.
These methods measure the similarity of two images in the pre-trained high-level vision networks, usually VGG classification network~\cite{simonyan_very_2015} trained in ImageNet~\cite{deng_imagenet_2009}.
Different perceptual objectives have been proposed in this category, \emph{e.g.}, $\mathcal{L}_{MSE}$ of features~\cite{dosovitskiy_generating_2016, johnson_perceptual_2016, ledig_photo-realistic_2017, zhang_unreasonable_2018, wang_esrgan_2018, mei2021sdan}, contextual objective~\cite{mechrez_contextual_2018, mechrez_maintaining_2018}, and semantic label~\cite{liu_connecting_2018, liu_connecting_2020}.
However, these methods lack a reasonable explanation of the effectiveness led by the perceptual objective~\cite{zhang_unreasonable_2018}.
Furthermore, the frozen network pre-trained on certain datasets, \emph{e.g.}, ImageNet, is not appropriate for the image restoration tasks conducted on the large-scale, diverse natural image datasets~\cite{agustsson_ntire_2017} or specific semantic image datasets~\cite{cordts_cityscapes_2016, liu_deep_2015, le_interactive_2012}.
Here we hypothesize that these issues come from the objectives that estimate the sample-to-sample distance in the feature space.
By exploiting the characteristics of single-semantic patches from natural images, which can be associated with an embedded manifold, we implicitly measure the divergence of the probability distributions estimated from restored images and clear images in the semantic feature space, and we use it as the objective to bypass the above issues.

Similar ideas that minimize the distribution divergence instead of the sample-to-sample distance have been proposed before.
In the area of image restoration, Contextual loss~\cite{mechrez_contextual_2018, mechrez_maintaining_2018} that proposed for misaligned image transformation implicitly minimizes the divergence between restored images and clear images.
It approximates the divergence by the contextual relationships within patches from a single image (\emph{i.e.} single image statistics~\cite{zontak_internal_2011}), and hence enables image-to-image translation to be conducted on the misaligned image pairs.
However, its performance is usually limited by the low accuracy of feature matching~\cite{zhang_zoom_2019} and leads to worse restoration performance in aligned image restoration learning.
More similar works that avail statistical features are GMMN~\cite{li2015generative} and GFMN~\cite{santos2019learning}.
They achieve the generative ability without adversarial learning in the problematic min/max game.
Nevertheless, they are not designed for the image restoration learning that majorly consists of natural images with diverse appearance, and the desired superiority cannot be gained here.
In the area of domain adaption, minimizing the statistics feature difference of the high-level vision networks can help networks adapt to unseen domain directly, like CORAL~\cite{sun_deep_2016} and MMD~\cite{tolstikhin2016minimax}.
However, these methods require semantic labels, which is not practical in the real-world image restoration datasets.
By exploiting the internal statistics~\cite{zontak_internal_2011} of natural images, our proposed method successfully facilitates the restoration learning through more accurate divergence approximation.

\begin{figure}[t]
  \centering
  \includegraphics[width=.5\linewidth]{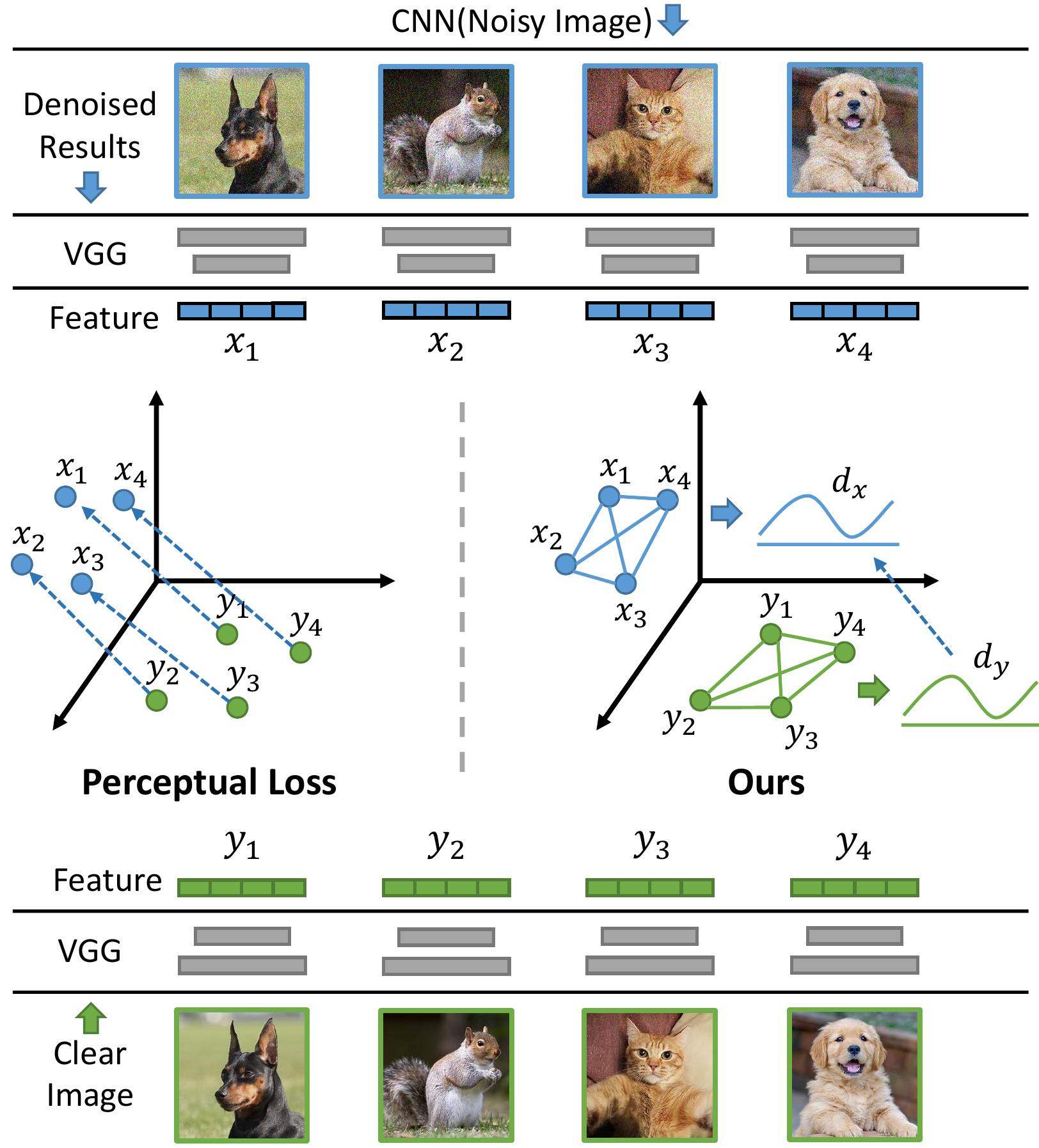}
  \caption{\textbf{Perceptual loss vs. Ours.} We minimize the distribution divergence between a set of restored images and the corresponding clear images, instead of the sample-to-sample distance, in the semantic feature space (\eg the penultimate layer of VGG). This procedure better simplifies the restoration learning and ameliorates underfitting compared with the perceptual loss.}
  \label{fig:teaser}
\end{figure}

\section{Method}
Let $\mathcal{X} \subset \mathbb{R}^{H\times W\times C}$ denotes the domain of degraded images caused by factors like noising, and $\mathcal{Y} \subset \mathbb{R}^{H\times W\times C}$ denotes the domain of corresponding clear images.
We wish to restore $x \in \mathcal{X}$ to appear like its corresponding target image $y \in \mathcal{Y}$ by using a denoising network $G(\cdot)$ that outputs $\tilde{y} = G(x)$.
To force the outputs $\tilde{y}$ maintains as much the perceptual detail as possible, recent works~\cite{dosovitskiy_generating_2016, johnson_perceptual_2016, ledig_photo-realistic_2017, mechrez_contextual_2018, mechrez_maintaining_2018} exploit the pre-trained high-level vision networks (\eg, the intermediate layer of VGG), denoted as $\Phi(\cdot)$, to guide the restoration learning by minimizing the similarity between $\tilde{y}$ and $y$ in the feature space of $\Phi(*)$.
This can be formulated as the objective with the similarity metric $D(\cdot)$:
\begin{equation}
  \mathcal{L}(x, y, G) = D(\Phi(y), \Phi(G(x))).
\end{equation}
In practical, the similarity metric $D(\cdot)$ is usually implemented by Mean Square Error (MSE) or Contextual Distance~\cite{mechrez_contextual_2018}.

Contrastively, we take the denoising learning as minimizing the divergence of probability distributions estimated by denoised images and clear images in the semantic feature space.
Given $N$ samples of image pairs that consist of $T_x = \{x_1, x_2, \dots, x_N\}$ and $T_y = \{y_1, y_2, \dots, y_N\}$, we incorporate the mutual information~\cite{torkkola_feature_2003} of them in the feature space of $\Phi(G(\cdot))$ into the restoration learning.
Such a manner is empirically proven to be effective to facilitate knowledge transferring~\cite{passalis_learning_2018, peng_correlation_2019, tung_similarity-preserving_2019, park_relational_2019, liu_knowledge_2019, chen_learning_2020, passalis_probabilistic_2020}.
By minimizing the divergence of the estimated probability distribution between samples $T_x$ in $\Phi(G(\cdot))$ and $T_y$ in $\Phi(\cdot)$, denoted as $\mathcal{G'}$ and $\mathcal{G}$, we force $G(\cdot)$ to better maintain the geometry of the feature space $\Phi(\cdot)$ estimated in the clear image domain $\mathcal{Y}$.
In doing so, we formulate the final objective as
\begin{equation}
  \mathcal{L}(T_x, T_y, G) = \sum^N_{i=1} \sum^N_{j=1, j \neq i} g'_{j|i} \log(\frac{g'_{j|i}}{g_{j|i}}).
\end{equation}
To elaborate, we will detail the divergence approximation in Section~\ref{sec:pkt}, and the sampling strategy in Section~\ref{sec:memo} and Section~\ref{sec:inner}.

\subsection{Probability Distribution Divergence}
\label{sec:pkt}
Here we model the correlation of samples from the same domain in the semantic feature space as the probability distribution.
Several methods have been proposed for modeling the correlation, including, but not limited to, probabilistic based~\cite{passalis_learning_2018, passalis_probabilistic_2020}, embedding based~\cite{peng_correlation_2019, chen_learning_2020}, graph based~\cite{liu_knowledge_2019}, and more~\cite{park_relational_2019}.
In this work, we exploit the kernel density estimation to estimate the probability distribution of samples in the semantic feature space, which describes the probability of each sample to select its neighbors~\cite{van_der_maaten_visualizing_2008}.
It is empirically proven to be effective for describing the geometry of feature space by Passalis et al.~\cite{passalis_learning_2018, passalis_probabilistic_2020}.
To elaborate, we denote the probability distribution between any two samples $i, j$ from the clear domain as $g_{i|j}$ and the restored domain as $g'_{i|j}$.
Based on the extracted feature $f^x$ and $f^y$ from $\Phi(G(\cdot)))$ and $\Phi(\cdot)$, the probability distributions are estimated by:
\begin{equation}
  g'_{i|j} = \frac{K_{cosine}(f_i^x, f_j^x)}{\sum^N_{k=1, k\neq j} K_{cosine}(f_k^x, f_j^x)} \in [0, 1],
  \label{eq:kernel}
\end{equation}
and
\begin{equation}
  g_{i|j} = \frac{K_{cosine}(f_i^y, f_j^y)}{\sum^N_{k=1, k\neq j} K_{cosine}(f_k^y, f_j^y)} \in [0, 1],
\end{equation}
where the cosine kernel function $K_{cosine}$ is employed for estimating the probabilty distribution, formulated with two vectors $a$ and $b$ as:
\begin{equation}
  K_{cosine} (a,b) = \frac{1}{2}(\frac{a^\top b}{||a||_2 ||b||_2}+1) \in [0, 1].
\end{equation}
As Turlach et al.~\cite{turlach_bandwidth_1999} suggested, this kernel function avoids the bandwidth choosing in Gaussian kernel, and it boosts performance compared with the Euclidean measures as Wang et al.~\cite{wang_visual_2014} suggested.

To minimize the difference of two estimated probability distributions, we avail the Kullback-Leibler (KL) divergence as the similarity metric, formulated as:
\begin{equation}
  D_{KL}(\mathcal{G'}||\mathcal{G}) = \int_{\mathbf{t}} \mathcal{G'}(\mathbf{t}_x) \log \frac{\mathcal{G'}(\mathbf{t}_x)}{\mathcal{G}(\mathbf{t}_y)} d\mathbf{t}.
\end{equation}
where $\mathbf{t}_x \in \mathcal{X}$ and $\mathbf{t}_y \in \mathcal{Y}$.
In practical implementation, we avail the mini-batch that consists of $N$ samples for approximation, aiming for acceleration in a parallel fashion.

\subsection{Memorized Historic Sampling}
\label{sec:memo}
Intuitively, the number of selected samples in a mini-batch should be as large as possible during training.
However, in practical implementation, increasing the number of samples is greatly limited by the GPU memory.
Such a limitation is more serious in the extracted semantic feature space, and hence greatly limits the effectiveness of our method.

\begin{figure}[t]
  \centering
  \includegraphics[width=.8\linewidth]{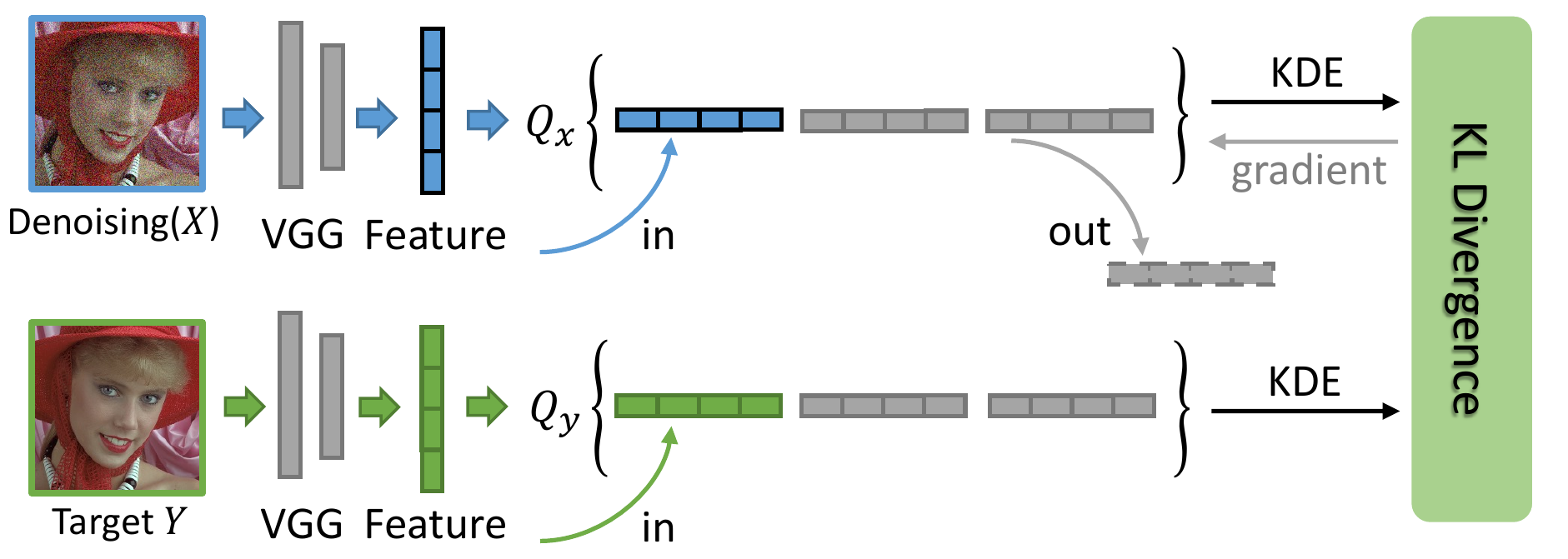}
  \caption{\textbf{Sampling with Historic Gradients.} We approximate the divergence with historic sampling by using two queues to bypass the GPU memory limits.}
  \label{fig:memory}
\end{figure}

To bypass the limitation, we introduce a Memorized Historic Sampling strategy, visualized in Figure~\ref{fig:memory}.
It maintains two \emph{queues} of feature samples, \emph{i.e.}, $Q^{\mathcal{X}}$ and $Q^{\mathcal{Y}}$ that can store historic features from previous mini-batches with limited GPU memory cost.
In doing so, we can estimate the probability distributions among queues instead of mini-batches.
Therefore, it allows a larger number of samples and a relatively smaller mini-batch used at runtime.
The queue is updated according to the First-In-First-Out rule, which enforces the historical samples in the queue are always newest, and hence it allows the probability distribution to be more consistent with the immediate state.
Based on such a strategy, we can formulate Equation~\ref{eq:kernel} as below:
\begin{equation}
  \label{eq:final}
  g'_{i|j} = \frac{K_{cosine}(Q^{\mathcal{X}}_i, Q^{\mathcal{X}}_j)}{\sum^q_{k=1, k\neq j} K_{cosine}(Q_k^{\mathcal{X}}, Q_j^{\mathcal{X}})} \in [0, 1],
\end{equation}
and 
\begin{equation}
  Q^{\mathcal{X}}_{1\dots N}, Q^{\mathcal{X}}_{N\dots q} \leftarrow f^x_{\{1 \dots N\}}, Q^{\mathcal{X}}_{\{1 \dots q-N\}}.
\end{equation}
where $q$ is the queue size of the appied queue for extracted features $f_x$ from a sigle mini-batch with the number of $N$, and $q \gg N$.

For example, the maximum size of a mini-batch can only be 32 in a single GPU card with a memory of 12GB, but the number of samples is usually set as 128 to ensure the accuracy of the estimation, which is not practical in a single GPU.
By using the queue in the size of 128, we can directly use the current mini-batch with features from 3 historical memorized mini-batch to perform an estimation with 128 samples, while without using additional $12\times3$ GB memory at running time.
It is because the queue that saves historical features costs less GPU memory compared with the procedure of feature extraction.
Similar strategies for enlarging the number of samples also exist, \emph{e.g}, memory bank~\cite{wu_unsupervised_2018} and momentum encoder~\cite{he_momentum_2020}.
Compared with them, our memorized historic queue is simpler but also enlarges the maximum number of samples to be used without additional GPU memory.
In the supplement we provide discussion about the effects of different queue sizes.

\subsection{Patch-Wise Internal Probabilities}
\label{sec:inner}

\begin{figure}[t]
  \centering
  \includegraphics[width=0.6\linewidth]{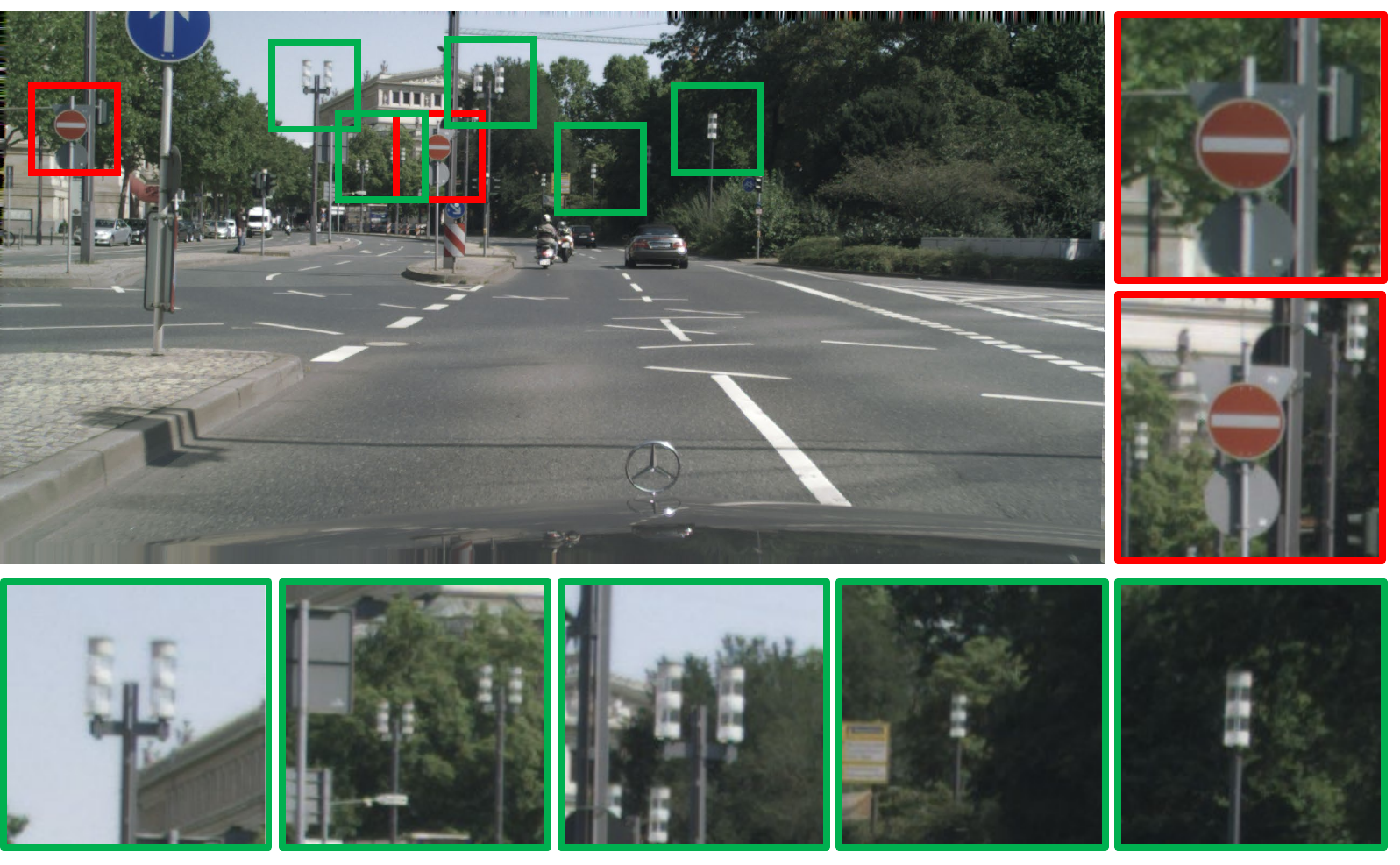}
  \caption{\textbf{Sampling with Internal Patches.}  Patches cropped from a single image may consist of different semantic objects that showed in different appearances.}
  \label{fig:internal}
\end{figure}

The most straightforward way to construct samples for the mini-batch is to randomly choose multiple images from the domains $\mathcal{X}$ and $\mathcal{Y}$, respectively.
Even though it is elegant, it fails to exploit another crucial probability in the single image, \emph{i.e.}, internal probability of patches from a single image, or internal statistics, which has been widely employed and empirically evaluated in many image restoration tasks~\cite{zontak_internal_2011, shocher_zero-shot_2018,shaham_singan_2019, park_contrastive_2020}.
As illustrated in Figure~\ref{fig:internal}, we can notice that the cropped patches from the single image, specifically the complex-semantic image, can be seen as multiple single-semantic images.
In addition, the probability estimated on multiple complex-semantic images is not always accurate in some cases, \emph{e.g.}, the denoising learning~\cite{zhang_ffdnet_2018} in mixed noise levels.
In such a case, conventional sampling results in an inaccurate estimation, because the restored images in the mini-batch come from different noisy levels.
These restored images distribute in different manifolds in the semantic feature space.
In contrast, by extracting patches from a single image resulted from extreme similar degradation, their similarity allows the probabilities to be accurately estimated.

In our method, a sliding window in a spatial size of $K \times K$ is availed to extract patches from the restored image $\tilde{Y}$ and clear image $Y$.
These patches are then inputted into the high-level vision network $\Phi(\cdot)$ as a single mini-batch and transformed into feature sets, which can be formulated as:
\begin{align}
  f^x = \Phi(\mathrm{sliding\_window} (\tilde{Y})).
\end{align}
Then we exploit Equation~\ref{eq:final} to estimate the probability distribution in the feature sets respectively for divergence approximation.
In the empirical evaluation, shown in Table~\ref{tab:denoising}, though the vanilla version achieves gains on the restoration performance, the generated images contain less semantic details (lower MIoU performance).
This decreasing may boil down to incorrect probability estimation.
In contrast, with the help of internal probability, our restoration network achieves a performance leap in both the restoration and semantic evaluation.

\section{Experiments}
Different from conventional denoising works, we focus on not only the restoration performance, but also the semantic accuracy, \ie, how the denoised images can be understood by semantic segmentation networks, as well as whether the method can be extended into the other restoration tasks.
Such a similar evaluation protocol is also employed in recent restoration works~\cite{li_aod-net_2017, li_mdcn_2020, tian_contrastive_2019}.
Therefore, our experiments are divided into three parts, including \textit{Cityscape Denoising and Segmentation}~\cite{cordts_cityscapes_2016}, \textit{Face Super-resolution and Alignment}~\cite{liu_deep_2015}, and \textit{Natural Image Restoration}~\cite{ancuti_i-haze_2018}.
More details please refer to the supplemental.

\subsection{Cityscape Denoising and Segmentation}
\label{sec:exp}
To demonstrate the superiority of our method, we conduct complementary denoising and segmentation experiments on the Cityscapes dataset.
The most representative denoising network, \ie, FFDNet~\cite{zhang_ffdnet_2018}, as well as the state-of-the-art denoising networks, \ie, CBDNet~\cite{guo_toward_2019} and SADNet~\cite{chang_spatial-adaptive_2020} are availed as the generation network $G(\cdot)$.
Various objectives are applied, \ie, $\mathcal{L}_1$, $\mathcal{L}_{SSIM}$~\cite{wang_image_2004}, $\mathcal{L}_{Perceptual}$~\cite{johnson_perceptual_2016}, $\mathcal{L}_{LPIPS}$~\cite{zhang_unreasonable_2018}, $\mathcal{L}_{Contextual}$~\cite{mechrez_contextual_2018}, $\mathcal{L}_{CrossEntropy}$~\cite{liu_connecting_2020}, and ours.
Notably, $\mathcal{L}_{CrossEntropy}$ is conducted with the HRNet48 that pre-trained on the Cityscapes dataset, which requires semantic labels during the denoising learning.
In contrast, ours does not need any additional data.
We then modify the original loss function of CBDNet and SADNet, \ie, $\mathcal{L}_2$ + $\mathcal{L}_{Asymmetric}$ + $\mathcal{L}_{TV}$ and $\mathcal{L}_2$, by attaching our proposed objective.
For convenience, we set the size of the sliding window $K$ as 224 and its stride as 56.

\begin{table*}[htbp]
    \centering
    \vspace{-10px}
    \caption{\textbf{Quantitative performance comparison on the cityscape denoising and segmentation.} The comprasion is conducted with various state-of-the-art denoising objectives and ours on the representative denoising networks.}
    \label{tab:denoising}
    \resizebox{\linewidth}{!}{
      \begin{tabular}{llccccccccc}
        \toprule
                                                                           &                                                                                & \multicolumn{3}{c}{Noise-Level $\sigma$=25} & \multicolumn{3}{c}{Noise-Level $\sigma$=35} & \multicolumn{3}{c}{Noise-Level $\sigma$=50}                                                                                                                                                         \\
        \cmidrule(r){3-5} \cmidrule(r){6-8} \cmidrule(r){9-11}
        Method (Backbone)                                                  & Objective                                                                      & PSNR $\uparrow$                             & SSIM $\uparrow$                             & MIoU (\%) $\uparrow$                        & PSNR $\uparrow$         & SSIM $\uparrow$        & MIoU (\%) $\uparrow$   & PSNR $\uparrow$         & SSIM $\uparrow$        & MIoU (\%) $\uparrow$   \\
        \midrule[1pt]
        \multirow{6}{*}{FFDNet~\cite{zhang_ffdnet_2018}} & $\mathcal{L}_1$                                                                & 35.033$^{(6)}$                              & 0.925$^{(6)}$                               & 0.605$^{(8)}$                               & 34.074$^{(6)}$          & 0.912$^{(6)}$          & 0.537$^{(8)}$          & 32.845$^{(6)}$          & 0.895$^{(6)}$          & 0.451$^{(7)}$          \\
                                                                           & + $\mathcal{L}_{SSIM}$~\cite{wang_image_2004}                & 35.567$^{(3)}$                              & 0.935$^{(2)}$                               & 0.642$^{(2)}$                               & 34.469$^{(4)}$          & 0.922$^{(2)}$          & 0.584$^{(2)}$          & 33.180$^{(3)}$          & 0.906$^{(2)}$          & 0.450$^{(8)}$          \\
                                                                           & + $\mathcal{L}_{Perceptual}$~\cite{johnson_perceptual_2016}  & 34.319$^{(7)}$                              & 0.912$^{(7)}$                               & 0.629$^{(4)}$                               & 33.486$^{(7)}$          & 0.899$^{(7)}$          & 0.582$^{(4)}$          & 32.383$^{(7)}$          & 0.881$^{(7)}$          & 0.509$^{(2)}$          \\
                                                                           & + $\mathcal{L}_{LPIPS}$~\cite{zhang_unreasonable_2018}       & 35.551$^{(4)}$                              & 0.929$^{(4)}$                               & 0.613$^{(6)}$                               & 34.463$^{(5)}$          & 0.916$^{(4)}$          & 0.541$^{(7)}$          & 33.138$^{(5)}$          & 0.899$^{(4)}$          & 0.452$^{(6)}$          \\
                                                                           & + $\mathcal{L}_{Contextual}$~\cite{mechrez_contextual_2018}  & 25.115$^{(8)}$                              & 0.762$^{(8)}$                               & 0.628$^{(5)}$                               & 24.938$^{(8)}$          & 0.758$^{(8)}$          & 0.583$^{(3)}$          & 24.775$^{(8)}$          & 0.753$^{(8)}$          & 0.509$^{(2)}$          \\
                                                                           & + $\mathcal{L}_{CrossEntropy}$~\cite{liu_connecting_2020}    & 35.913$^{(2)}$                              & 0.932$^{(3)}$                               & 0.630$^{(3)}$                               & 34.800$^{(2)}$          & 0.919$^{(3)}$          & 0.565$^{(5)}$          & 33.477$^{(2)}$          & 0.903$^{(3)}$          & 0.491$^{(4)}$          \\
        \hline\multirow{2}{*}{D2SM (Ours)}                                 & w/o. Internal                                                         & 35.543$^{(5)}$                              & 0.929$^{(4)}$                               & 0.612$^{(7)}$                               & 34.475$^{(3)}$          & 0.916$^{(4)}$          & 0.546$^{(6)}$          & 33.167$^{(4)}$          & 0.899$^{(4)}$          & 0.463$^{(5)}$          \\
                                                                           & w/. Internal                                                  & \textbf{36.454$^{(1)}$}                     & \textbf{0.936$^{(1)}$}                      & \textbf{0.644$^{(1)}$}                      & \textbf{35.206$^{(1)}$} & \textbf{0.923$^{(1)}$} & \textbf{0.587$^{(1)}$} & \textbf{33.807$^{(1)}$} & \textbf{0.907$^{(1)}$} & \textbf{0.520$^{(1)}$} \\
  
        \hline
        \hline
        \multirow{3}{*}{CBDNet~\cite{guo_toward_2019}}                   &  -             & 36.152$^{(3)}$                              & 0.936$^{(2)}$                               & 0.655$^{(3)}$                               & 34.964$^{(3)}$          & 0.923$^{(3)}$          & 0.599$^{(3)}$          & 33.613$^{(3)}$          & 0.907$^{(3)}$          & 0.539$^{(3)}$          \\
                                                       & w/o. Internal                    & 36.254$^{(2)}$                              & 0.935$^{(3)}$                               & 0.679$^{(2)}$                               & 35.158$^{(2)}$          & 0.925$^{(2)}$          & 0.631$^{(2)}$          & 33.904$^{(2)}$          & 0.911$^{(2)}$          & 0.550$^{(2)}$          \\
                                                       & w/. Internal                     & \textbf{36.899$^{(1)}$}                     & \textbf{0.941$^{(1)}$}                      & \textbf{0.691$^{(1)}$}                      & \textbf{35.596$^{(1)}$} & \textbf{0.929$^{(1)}$} & \textbf{0.652$^{(1)}$} & \textbf{34.172$^{(1)}$} & \textbf{0.914$^{(1)}$} & \textbf{0.600$^{(1)}$} \\
        \hline
        \hline
        \multirow{3}{*}{SADNet~\cite{chang_spatial-adaptive_2020}}        & -     & 36.310$^{(3)}$                              & 0.936$^{(3)}$                               & 0.674$^{(3)}$                               & 35.081$^{(3)}$          & 0.924$^{(2)}$          & 0.637$^{(3)}$         & 33.730$^{(3)}$          & 0.908$^{(3)}$          & 0.581$^{(3)}$    \\
        & w/o. Internal                                            & 36.822$^{(2)}$                               & 0.940$^{(2)}$                                         & 0.691$^{(2)}$                                & 35.247$^{(2)}$             & 0.924$^{(2)}$          & 0.655$^{(2)}$         & 34.133$^{(2)}$          & 0.912$^{(2)}$          & 0.600$^{(2)}$    \\
        & w/. Internal                                            & \textbf{37.130$^{(1)}$}                     & \textbf{0.943$^{(1)}$}                      & \textbf{0.701$^{(1)}$}                      & \textbf{35.839$^{(1)}$} & \textbf{0.931$^{(1)}$} & \textbf{0.670$^{(1)}$} & \textbf{34.440$^{(1)}$} & \textbf{0.916$^{(1)}$} & \textbf{0.634$^{(1)}$} \\
        \bottomrule
      \end{tabular}}
      \vspace{-10px}
  \end{table*}

For denoising training, we construct noisy images by adding additive color Gaussian noise of noise level $\sigma \in [0, 75]$ to the clean images from the Cityscapes training set.
The images are randomly cropped into $512\times 512$ patches in a mini-batch size of 64.
Other settings are kept the same as the settings in FFDNet.
For evaluation, we first measure appearance similarities between restored images and corresponding clear images in the Cityscapes validation set, in noisy levels $\{25, 35, 50\}$, which is commonly selected by the denoising community.
We then measure the semantic segmentation accuracy on restored images in the term of Mean Intersection-over-Union (MIoU) in 19 pre-defined semantic classes, i.e., \emph{road}, \emph{sidewalk}, \emph{building}, \emph{wall}, \emph{fence}, \emph{pole}, \emph{traffic light}, \emph{traffic sign}, \emph{vegetation}, \emph{terrain}, \emph{sky}, \emph{person}, \emph{rider}, \emph{car}, \emph{truck}, \emph{bus}, \emph{train}, \emph{motorcycle}, and \emph{bicycle}.

\noindent \textbf{Quantitative Comparison.}
In Table~\ref{tab:denoising}, it is easy to see that ours outperforms all compared objectives largely in PSNR, SSIM, and MIoU metrics on all noisy levels, when applied in the same backbone.
Compared with the state-of-the-art objectives that combine high-level vision tasks, \ie, $\mathcal{L}_{CrossEntropy}$~\cite{liu_connecting_2020}, which requires the semantic label of images during training, ours still outperforms it by 0.542dB in PSNR, 1.4\% in MIoU, without using any additional data.
Besides, ours shows strong robustness when adopted with different network architectures and objectives, \eg, it helps the original CBDNet improves 0.747dB in PSNR and 0.5\% in MIoU.
Also, the comparison between using and without using internal probability further demonstrates its superiority for complex-semantic images.

\begin{figure*}[htbp]
    \begin{subfigure}[t]{.195\linewidth}
      \captionsetup{justification=centering, labelformat=empty, font=scriptsize}
      \includegraphics[width=1.\linewidth]{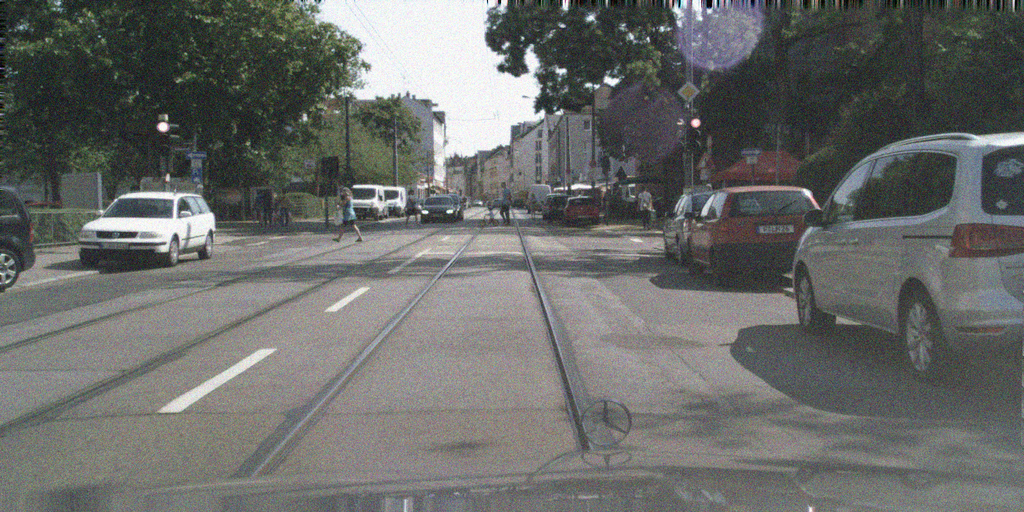}
      \includegraphics[width=1.\linewidth]{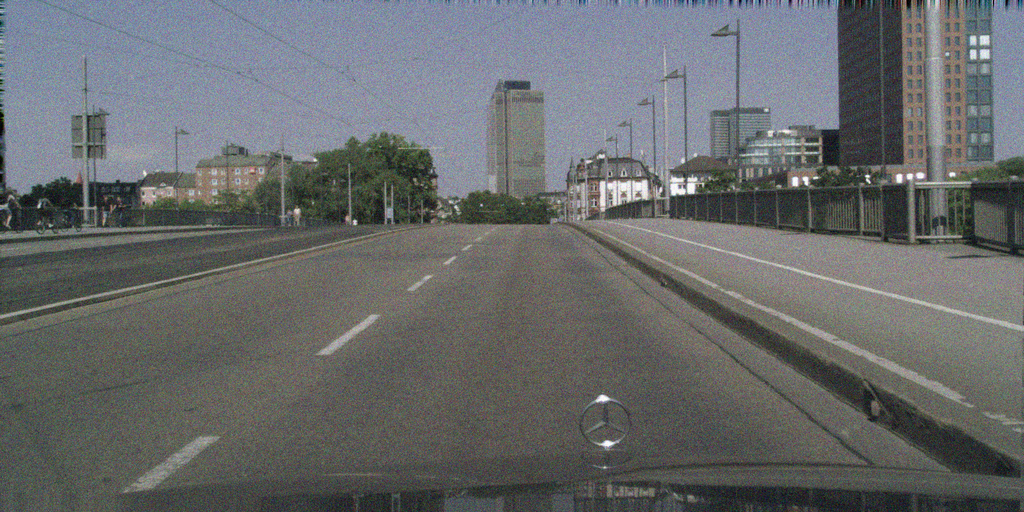}
      \caption{Noisy Image}
    \end{subfigure}%
    \hfill
    \begin{subfigure}[t]{.195\linewidth}
      \captionsetup{justification=centering, labelformat=empty, font=scriptsize}
      \includegraphics[width=1.\linewidth]{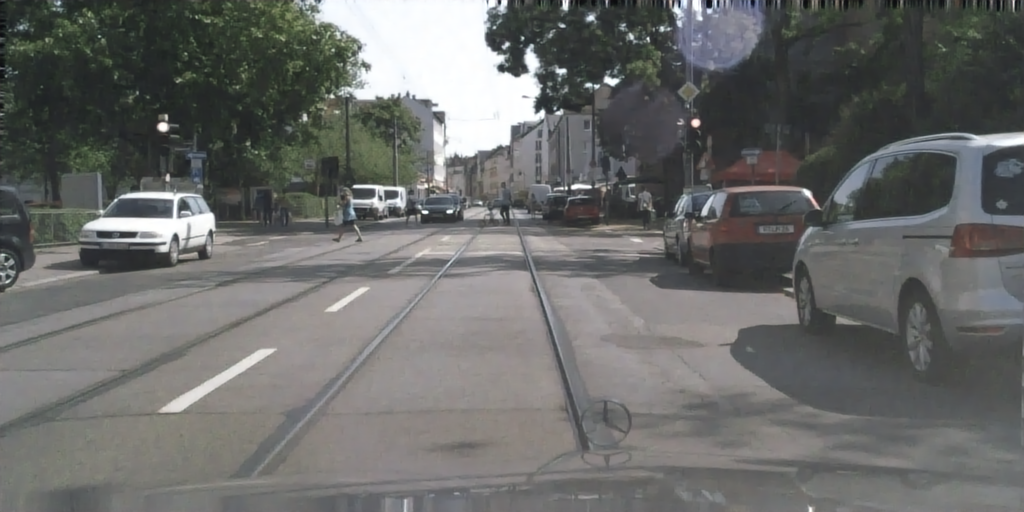}
      \includegraphics[width=1.\linewidth]{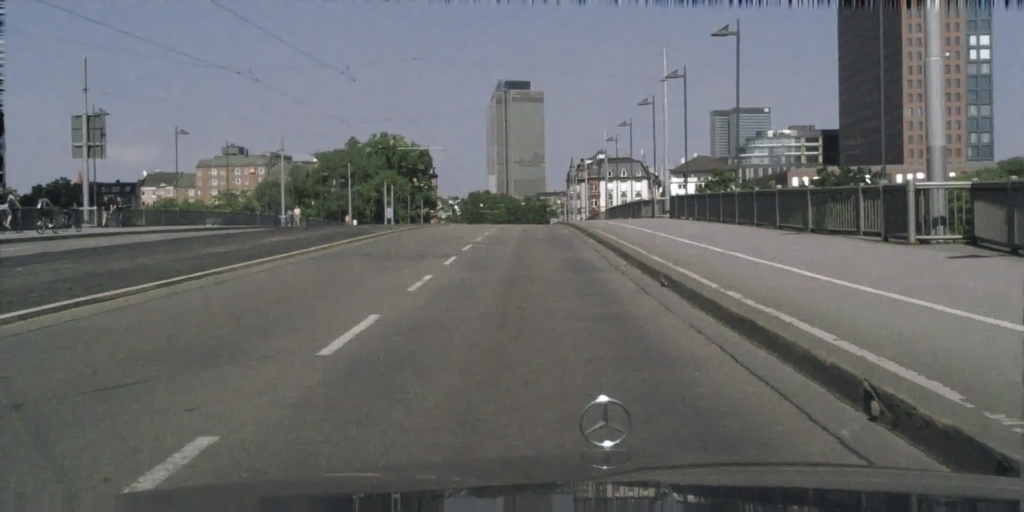}
      \caption{$\mathcal{L}_1$}
    \end{subfigure}%
    \hfill
    \begin{subfigure}[t]{.195\linewidth}
      \captionsetup{justification=centering, labelformat=empty, font=scriptsize}
      \includegraphics[width=1.\linewidth]{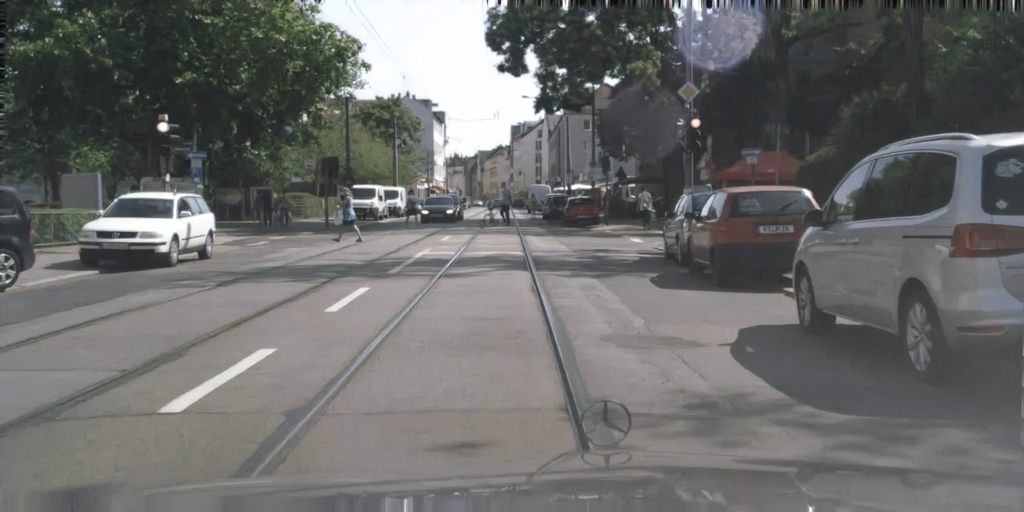}
      \includegraphics[width=1.\linewidth]{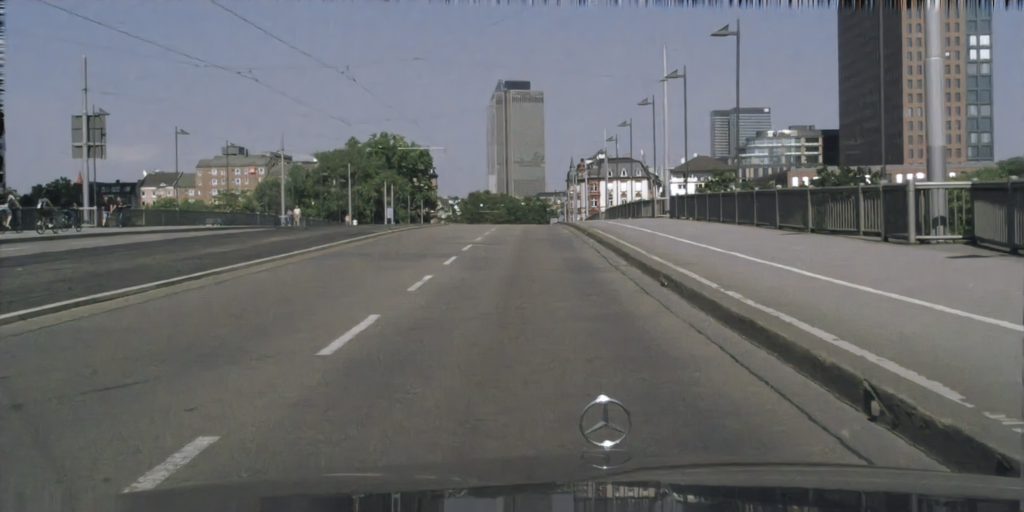}
      \caption{$\mathcal{L}_{SSIM}$~\cite{wang_image_2004}}
    \end{subfigure}%
    \hfill
    \begin{subfigure}[t]{.195\linewidth}
      \captionsetup{justification=centering, labelformat=empty, font=scriptsize}
      \includegraphics[width=1.\linewidth]{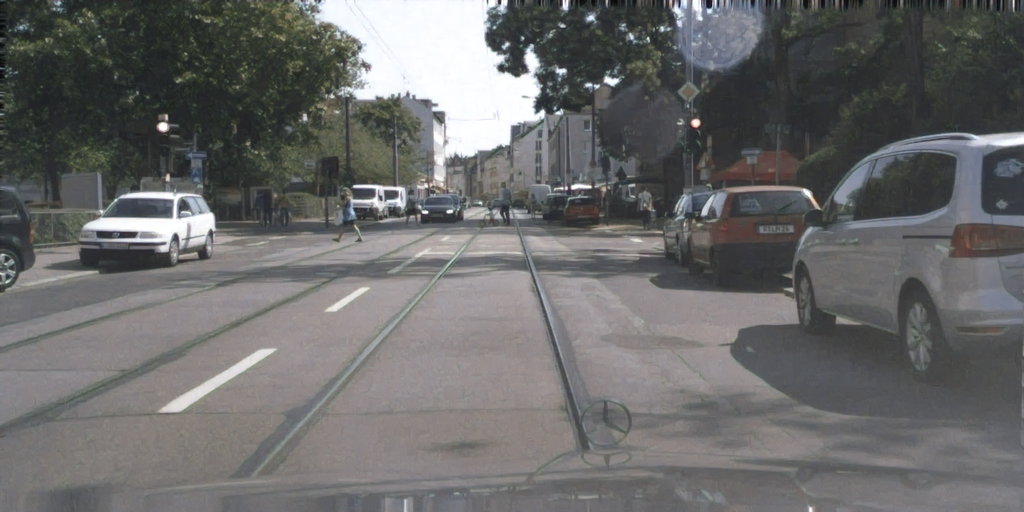}
      \includegraphics[width=1.\linewidth]{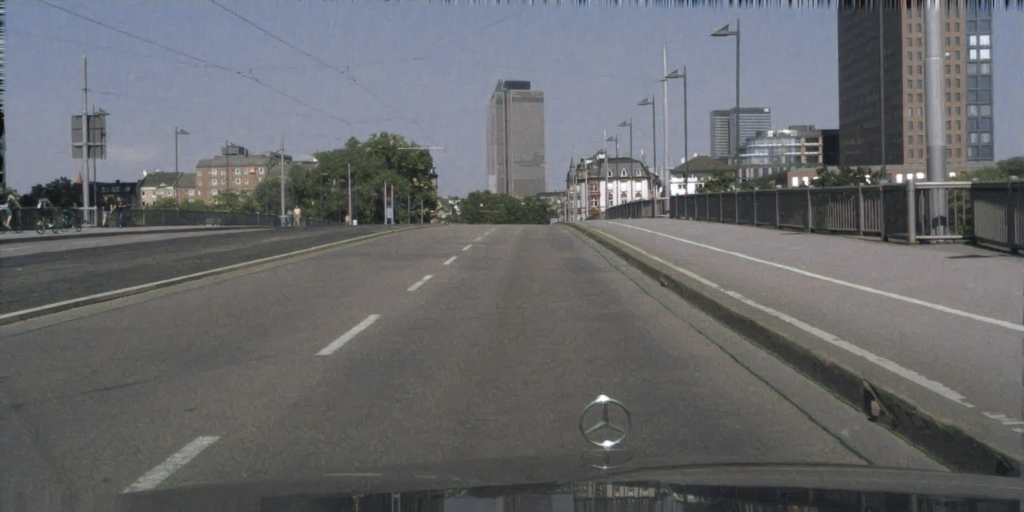}
      \caption{$\mathcal{L}_{Perceptual}$~\cite{johnson_perceptual_2016}}
    \end{subfigure}%
    \hfill
    \begin{subfigure}[t]{.195\linewidth}
      \captionsetup{justification=centering, labelformat=empty, font=scriptsize}
      \includegraphics[width=1.\linewidth]{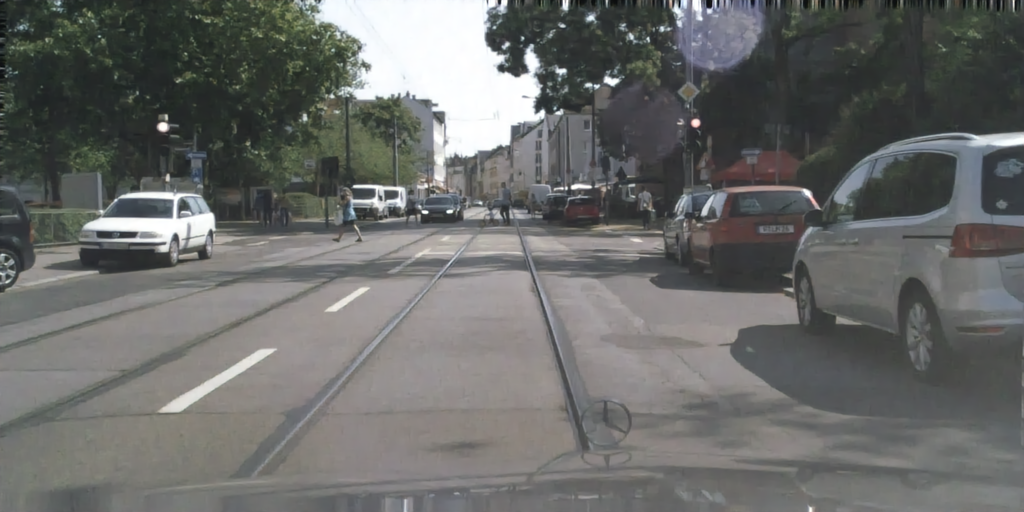}
      \includegraphics[width=1.\linewidth]{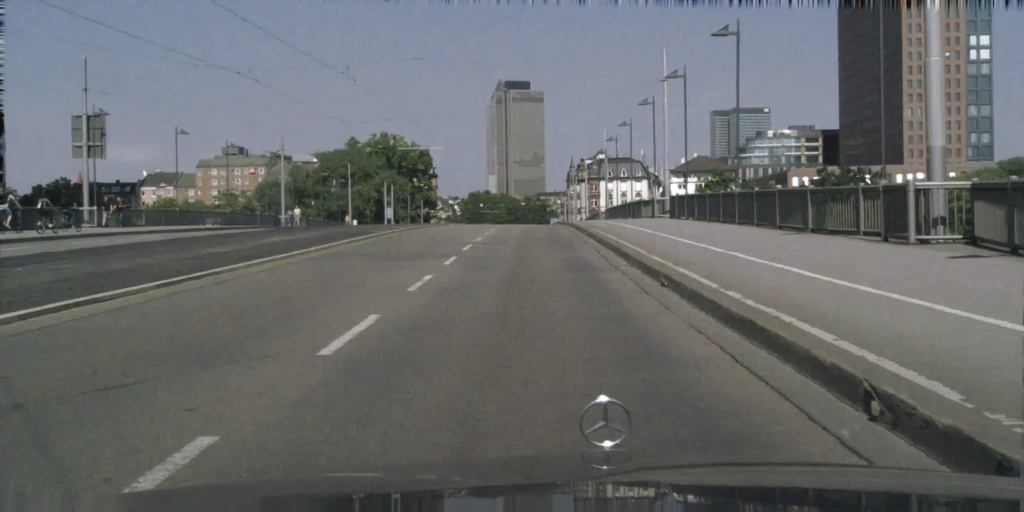}
      \caption{$\mathcal{L}_{LPIPS}$~\cite{zhang_unreasonable_2018}}
    \end{subfigure}%
  
    \begin{subfigure}[t]{.195\linewidth}
      \captionsetup{justification=centering, labelformat=empty, font=scriptsize}
      \includegraphics[width=1.\linewidth]{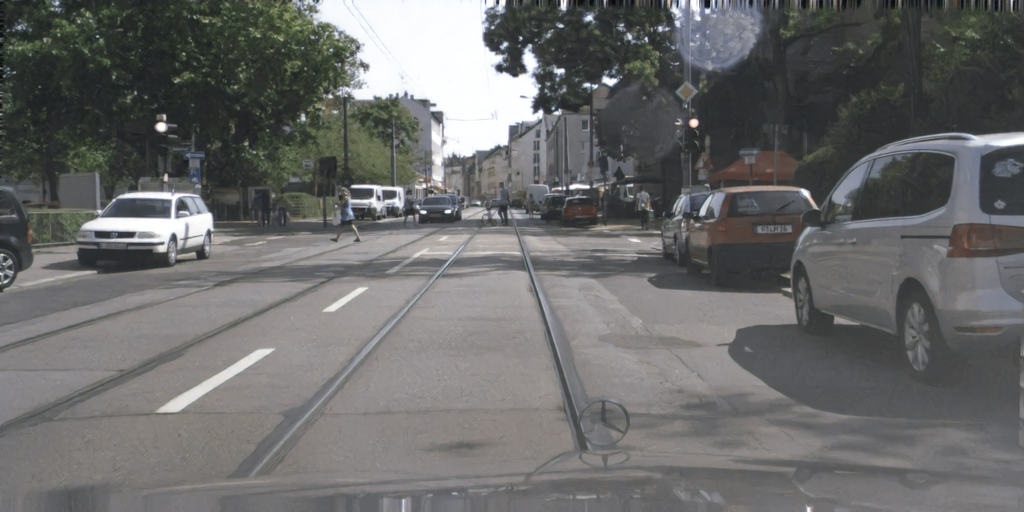}
      \includegraphics[width=1.\linewidth]{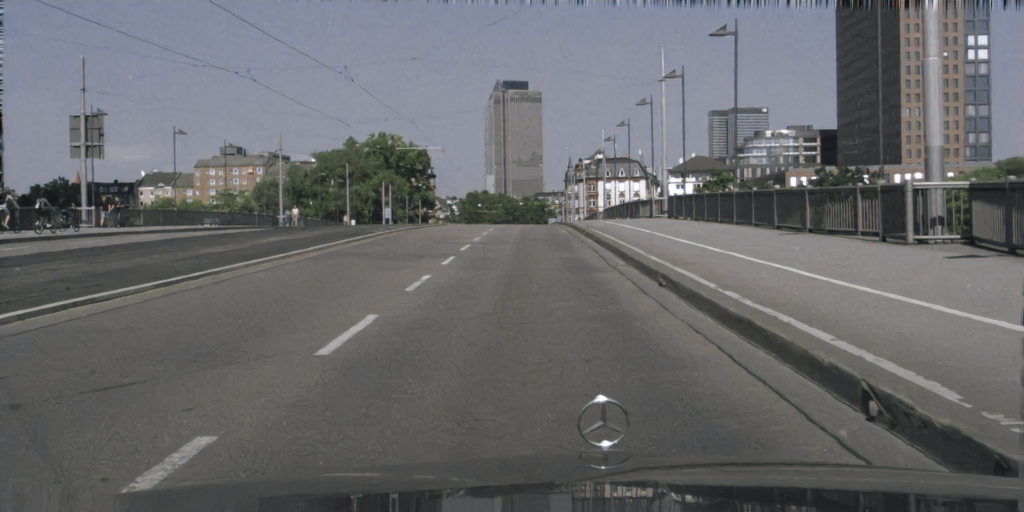}
      \caption{$\mathcal{L}_{Contextual}$~\cite{mechrez_contextual_2018}}
    \end{subfigure}%
    \hfill
    \begin{subfigure}[t]{.195\linewidth}
      \captionsetup{justification=centering, labelformat=empty, font=scriptsize}
      \includegraphics[width=1.\linewidth]{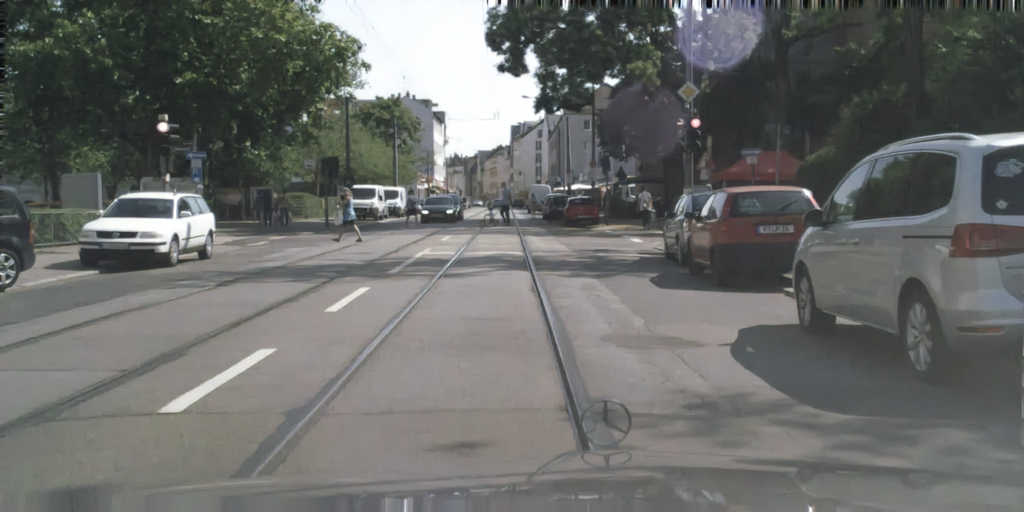}
      \includegraphics[width=1.\linewidth]{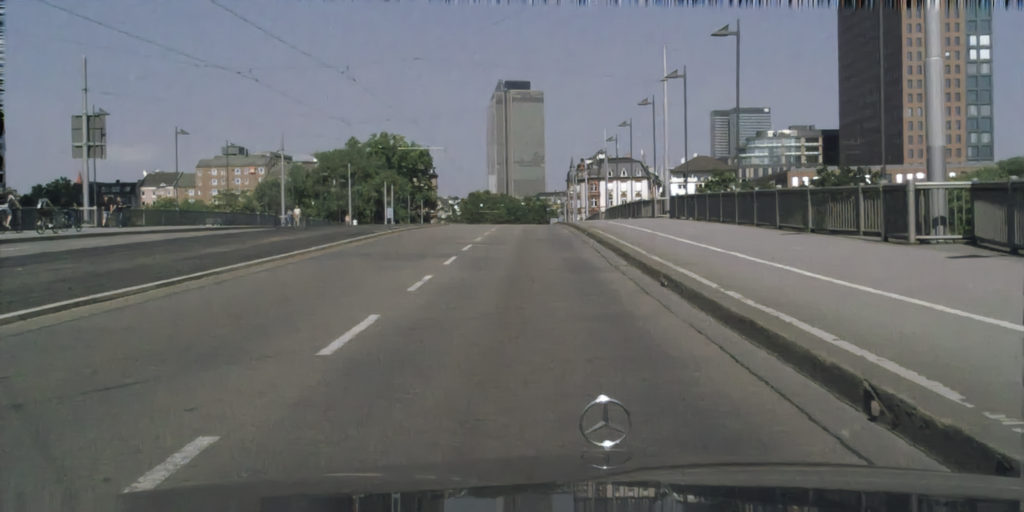}
      \caption{$\mathcal{L}_{CrossEntropy}$~\cite{liu_connecting_2020}}
    \end{subfigure}%
    \hfill
    \begin{subfigure}[t]{.195\linewidth}
      \captionsetup{justification=centering, labelformat=empty, font=scriptsize}
      \includegraphics[width=1.\linewidth]{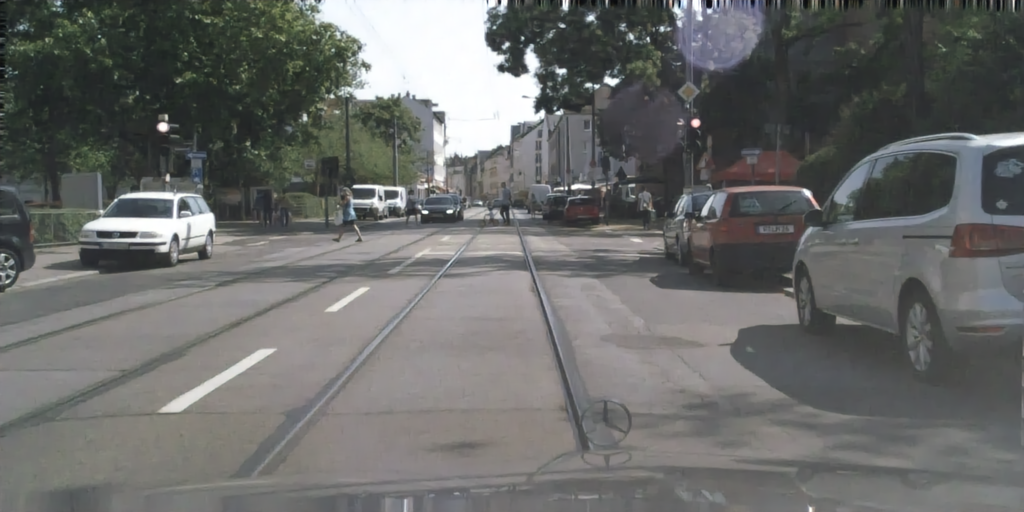}
      \includegraphics[width=1.\linewidth]{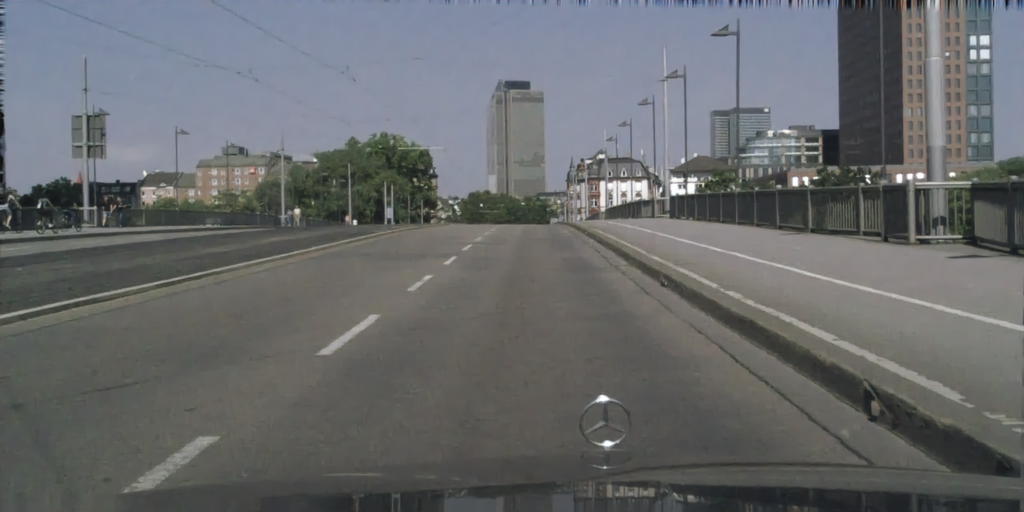}
      \caption{Ours w/o. Internal}
    \end{subfigure}%
    \hfill
    \begin{subfigure}[t]{.195\linewidth}
      \captionsetup{justification=centering, labelformat=empty, font=scriptsize}
      \includegraphics[width=1.\linewidth]{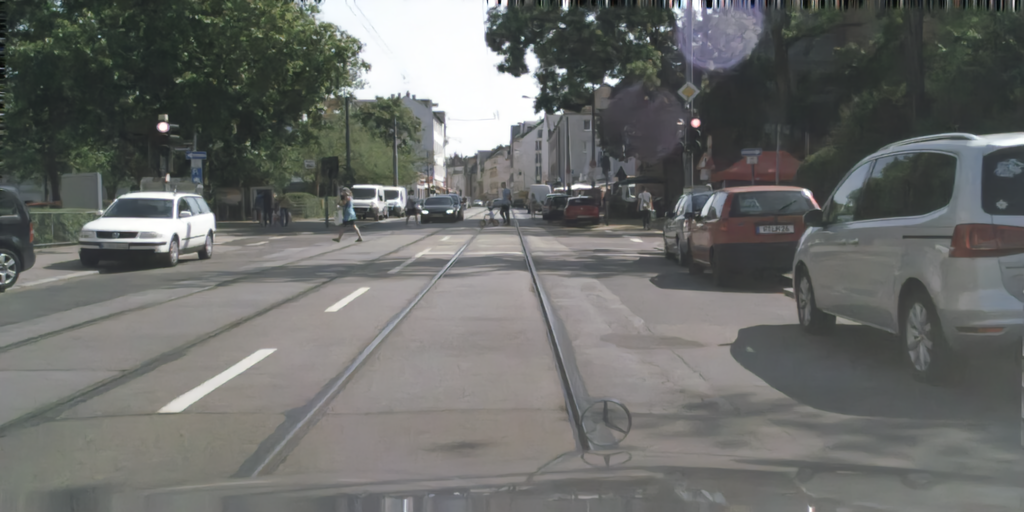}
      \includegraphics[width=1.\linewidth]{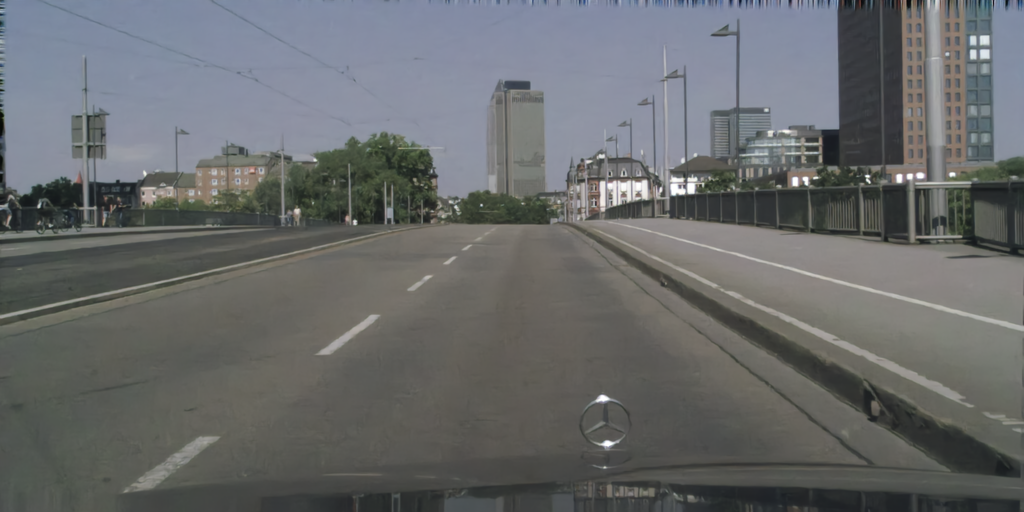}
      \caption{Ours}
    \end{subfigure}%
    \hfill
    \begin{subfigure}[t]{.195\linewidth}
      \captionsetup{justification=centering, labelformat=empty, font=scriptsize}
      \includegraphics[width=1.\linewidth]{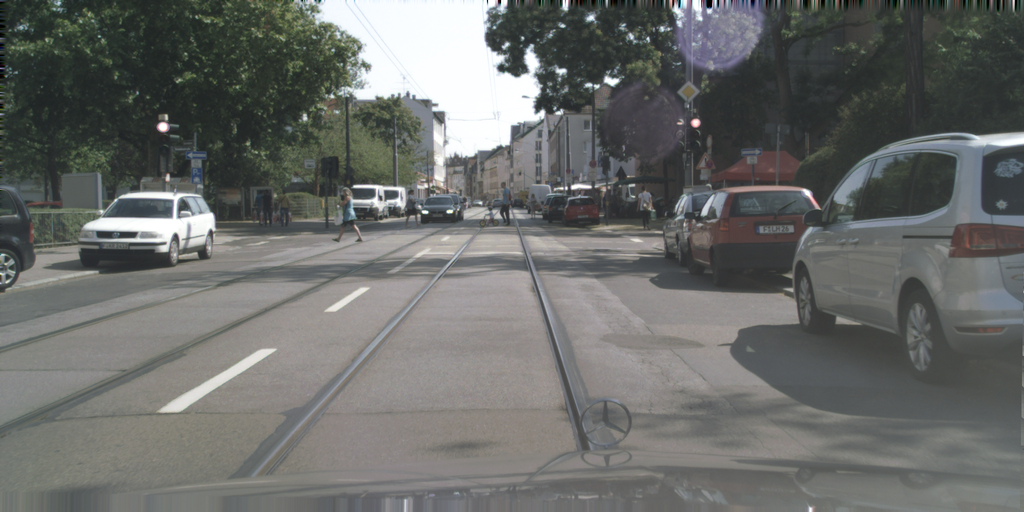}
      \includegraphics[width=1.\linewidth]{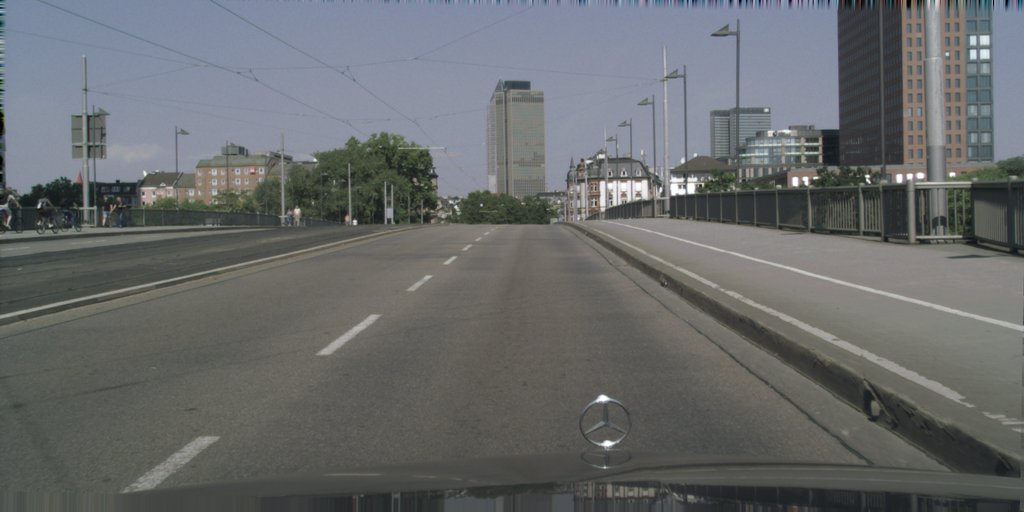}
      \caption{Clear Image}
    \end{subfigure}%
    \caption{Qualitative comparison on the denoising results. Ours results contain the most fine-grained high-frequency information and more visual pleasant details. (400\% Zoom is recommended to see their difference in details and color bias.)}
    \vspace{-10px}
    \label{fig:visual}
  \end{figure*}

\begin{figure*}[htbp]
    \begin{subfigure}[t]{.195\linewidth}
      \captionsetup{justification=centering, labelformat=empty, font=scriptsize}
      \setlength{\fboxrule}{1pt}
      \setlength{\fboxsep}{0pt}
      \fcolorbox{blue}{white}{\includegraphics[width=0.45\linewidth]{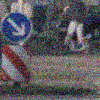}}
      \fcolorbox{red}{white}{\includegraphics[width=0.45\linewidth]{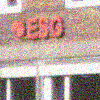}}
      \includegraphics[width=1.\linewidth]{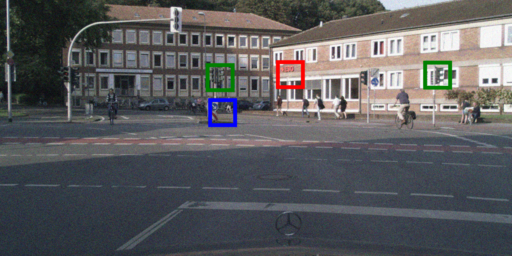}
      \caption{Noisy Image \\ Acc: 23.66\%}
    \end{subfigure}%
    \hfill
    \begin{subfigure}[t]{.195\linewidth}
      \captionsetup{justification=centering, labelformat=empty, font=scriptsize}
      \setlength{\fboxrule}{1pt}
      \setlength{\fboxsep}{0pt}
      \fcolorbox{blue}{white}{\includegraphics[width=0.45\linewidth]{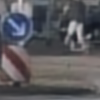}}
      \fcolorbox{red}{white}{\includegraphics[width=0.45\linewidth]{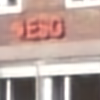}}
      \fcolorbox{green}{white}{\includegraphics[width=0.45\linewidth]{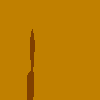}}
      \fcolorbox{green}{white}{\includegraphics[width=0.45\linewidth]{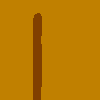}}
      \caption{$\mathcal{L}_1$ \\ Acc: 42.59\%}
    \end{subfigure}%
    \hfill
    \begin{subfigure}[t]{.195\linewidth}
      \captionsetup{justification=centering, labelformat=empty, font=scriptsize}
      \setlength{\fboxrule}{1pt}
      \setlength{\fboxsep}{0pt}
      \fcolorbox{blue}{white}{\includegraphics[width=0.45\linewidth]{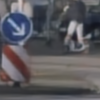}}
      \fcolorbox{red}{white}{\includegraphics[width=0.45\linewidth]{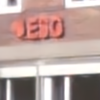}}
      \fcolorbox{green}{white}{\includegraphics[width=0.45\linewidth]{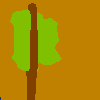}}
      \fcolorbox{green}{white}{\includegraphics[width=0.45\linewidth]{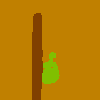}}
      \caption{$\mathcal{L}_{SSIM}$~\cite{wang_image_2004} \\ Acc: 44.84\%}
    \end{subfigure}%
    \hfill
    \begin{subfigure}[t]{.195\linewidth}
      \captionsetup{justification=centering, labelformat=empty, font=scriptsize}
      \setlength{\fboxrule}{1pt}
      \setlength{\fboxsep}{0pt}
      \fcolorbox{blue}{white}{\includegraphics[width=0.45\linewidth]{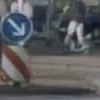}}
      \fcolorbox{red}{white}{\includegraphics[width=0.45\linewidth]{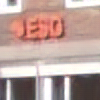}}
      \fcolorbox{green}{white}{\includegraphics[width=0.45\linewidth]{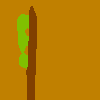}}
      \fcolorbox{green}{white}{\includegraphics[width=0.45\linewidth]{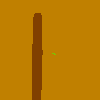}}
      \caption{$\mathcal{L}_{Perceptual}$~\cite{johnson_perceptual_2016} \\ Acc: 45.35\%}
    \end{subfigure}%
    \hfill
    \begin{subfigure}[t]{.195\linewidth}
      \captionsetup{justification=centering, labelformat=empty, font=scriptsize}
      \setlength{\fboxrule}{1pt}
      \setlength{\fboxsep}{0pt}
      \fcolorbox{blue}{white}{\includegraphics[width=0.45\linewidth]{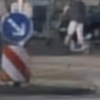}}
      \fcolorbox{red}{white}{\includegraphics[width=0.45\linewidth]{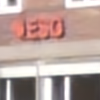}}
      \fcolorbox{green}{white}{\includegraphics[width=0.45\linewidth]{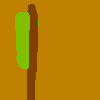}}
      \fcolorbox{green}{white}{\includegraphics[width=0.45\linewidth]{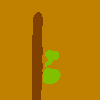}}
      \caption{$\mathcal{L}_{LPIPS}$~\cite{zhang_unreasonable_2018} \\ Acc: 43.25\%}
    \end{subfigure}%
  
    \begin{subfigure}[t]{.195\linewidth}
      \captionsetup{justification=centering, labelformat=empty, font=scriptsize}
      \setlength{\fboxrule}{1pt}
      \setlength{\fboxsep}{0pt}
      \fcolorbox{blue}{white}{\includegraphics[width=0.45\linewidth]{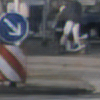}}
      \fcolorbox{red}{white}{\includegraphics[width=0.45\linewidth]{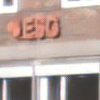}}
      \fcolorbox{green}{white}{\includegraphics[width=0.45\linewidth]{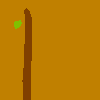}}
      \fcolorbox{green}{white}{\includegraphics[width=0.45\linewidth]{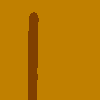}}
      \caption{$\mathcal{L}_{Contextual}$~\cite{mechrez_contextual_2018} \\ Acc: 43.28\%}
    \end{subfigure}%
    \hfill
    \begin{subfigure}[t]{.195\linewidth}
      \captionsetup{justification=centering, labelformat=empty, font=scriptsize}
      \setlength{\fboxrule}{1pt}
      \setlength{\fboxsep}{0pt}
      \fcolorbox{blue}{white}{\includegraphics[width=0.45\linewidth]{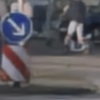}}
      \fcolorbox{red}{white}{\includegraphics[width=0.45\linewidth]{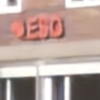}}
      \fcolorbox{green}{white}{\includegraphics[width=0.45\linewidth]{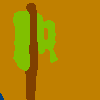}}
      \fcolorbox{green}{white}{\includegraphics[width=0.45\linewidth]{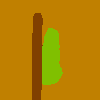}}
      \caption{$\mathcal{L}_{CrossEntropy}$~\cite{liu_connecting_2020} \\ Acc: 44.09\%}
    \end{subfigure}%
    \hfill
    \begin{subfigure}[t]{.195\linewidth}
      \captionsetup{justification=centering, labelformat=empty, font=scriptsize}
      \setlength{\fboxrule}{1pt}
      \setlength{\fboxsep}{0pt}
      \fcolorbox{blue}{white}{\includegraphics[width=0.45\linewidth]{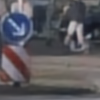}}
      \fcolorbox{red}{white}{\includegraphics[width=0.45\linewidth]{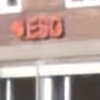}}
      \fcolorbox{green}{white}{\includegraphics[width=0.45\linewidth]{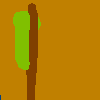}}
      \fcolorbox{green}{white}{\includegraphics[width=0.45\linewidth]{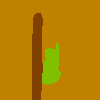}}
      \caption{Ours w/o. Internal \\ Acc: 42.90\%}
    \end{subfigure}%
    \hfill
    \begin{subfigure}[t]{.195\linewidth}
      \captionsetup{justification=centering, labelformat=empty, font=scriptsize}
      \setlength{\fboxrule}{1pt}
      \setlength{\fboxsep}{0pt}
      \fcolorbox{blue}{white}{\includegraphics[width=0.45\linewidth]{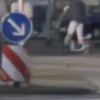}}
      \fcolorbox{red}{white}{\includegraphics[width=0.45\linewidth]{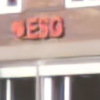}}
      \fcolorbox{green}{white}{\includegraphics[width=0.45\linewidth]{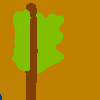}}
      \fcolorbox{green}{white}{\includegraphics[width=0.45\linewidth]{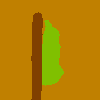}}
      \caption{Ours \\ Acc: 46.31\%}
    \end{subfigure}%
    \hfill
    \begin{subfigure}[t]{.195\linewidth}
      \captionsetup{justification=centering, labelformat=empty, font=scriptsize}
      \setlength{\fboxrule}{1pt}
      \setlength{\fboxsep}{0pt}
      \fcolorbox{blue}{white}{\includegraphics[width=0.45\linewidth]{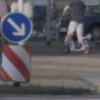}}
      \fcolorbox{red}{white}{\includegraphics[width=0.45\linewidth]{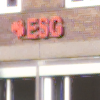}}
      \fcolorbox{green}{white}{\includegraphics[width=0.45\linewidth]{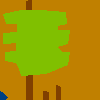}}
      \fcolorbox{green}{white}{\includegraphics[width=0.45\linewidth]{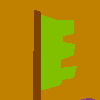}}
      \caption{Clear Image \\ Acc: 55.60\%}
    \end{subfigure}%
    \caption{Qualitative comparison on the denoising and segmentation results. Ours preserves most of the semantic details, including the human shape and font edge in the highlighted area. Additionally, in the shown segmentation results, our result is the only one that can be successfully recognized into \emph{traffic light}.}
    \label{fig:visualization}
    \vspace{-10px}
  \end{figure*}

\noindent \textbf{Qualitative Comparison.}
Though $\mathcal{L}_{Perceptual}$ applied in restoration methods has been proven to lead to better perceptual quality in restored images, we find that ours significantly outperforms it with more visually pleasant and exact details as shwon in Figure~\ref{fig:visual}.
As shown in Figure~\ref{fig:visualization}, our restored results best preserve the edge of the character ``\textbf{S}'' in the red rectangle area.
Besides, the blue rectangle area shows the best sharp details in our restored results compared with others.
With regard to the segmentation evaluation, restored results from restoration networks trained with ours can best be  segmented accurately.
For instance, as two green rectangle areas are shown in the Figure~\ref{fig:visualization}, our result is the only one that is successfully recognized into \emph{traffic light}.
This indicates that ours can best preserve semantic details during restoration in the way of divergence minimization.

\noindent \textbf{Distribution Visualization.} In order to get insights into the probability distribution, here we visualize the semantic feature space estimated by restored images and clear images.
To elaborate, we randomly select 500 animal images that belong to 10 categories, \ie, \emph{cat}, \emph{dog}, \emph{chicken}, \emph{cow}, \emph{horse}, \emph{sheep}, \emph{squirrel}, \emph{elephant}, \emph{butterfly}, and \emph{spider}.
We then process their noisy version (\ie adding additive color Gaussian noise of noise level $\sigma$=25) with the FFDNet pre-trained on the noisy Cityscapes dataset.
After that, the denoised images, as well as the clear images, are inputted into the pre-trained ResNet101~\cite{he_deep_2016} to extract semantic feature maps.
As such, we can visualize the distribution of semantic features with the t-SNE~\cite{van_der_maaten_visualizing_2008} in 2D coordinates.
Compared with others, the visualized distribution from our restored images best preserves the distribution of clear images in the semantic feature space.
This indicates that our proposed method indeed implicitly minimizes the probability distribution divergence between restored images and clear images in the semantic feature space.

\subsection{Face Super-resolution and Alignment}
Here we extend D2SM with historic sampling as the objective and conduct face super-resolution learning under the settings of DICNet~\cite{ma_deep_2020}.
For the evaluation of high-level vision applications, we exploit the face alignment as a measurement, and its accuracy is denoted in the term of NRMSE.
The queue size is set as 256 and the mini-batch size is 32, which means the probability is estimated among 32 face images instead interal patches.

In table~\ref{tab:facesr}, we show the quantitative performance comparison between DICNet and ours, as well as the other state-of-the-art methods.
Notably, the face alignment network is pre-trained on the CelebA dataset, and hence its evaluation on the Helen dataset is generally not good enough, which can only be used for reference.
By combining our objective with existing $\mathcal{L}_1$ and $\mathcal{L}_{Alignment}$ proposed by DICNet, our modified version successfully outperforms the original DICNet and DICNet in the GAN manner (DICGAN), in both the distortion measurement and alignment measurement.
In contrast, the original DICNet can only achieve leading performance in the distortion measurement but is poor in alignment measurement, while the DICGAN can only achieve leading performance in the alignment measurement but is bad at distortion measurement.

\begin{table*}[t]
    \caption{Quantitative performance comparison on the face super-resolution and alignment. By simply attaching our objective into the DICNet, our method can outperform the state-of-the-art DICNet and DICGAN in both the distortion measurement and high-level vision application measurement.}
    \label{tab:facesr}
    \resizebox{\linewidth}{!}{
      \begin{tabular}{llcccccc}
        \toprule
                                                                           &                                                                                                                          & \multicolumn{3}{c}{CelebA Dataset} & \multicolumn{3}{c}{Helen Dataset}                                                                                                         \\
        \cmidrule(r){3-5} \cmidrule(r){6-8}
        Method (Backbone)                                                  & Objective                                                                                                                & PSNR $\uparrow$                    & SSIM $\uparrow$                   & NRMSE $\downarrow$      & PSNR $\uparrow$        & SSIM $\uparrow$         & NRMSE $\downarrow$       \\
        \hline \hline
        Bicubic                                                            & -                                                                                                                        & 23.58$^{(10)}$                     & 0.6285$^{(10)}$                   & 0.3385$^{(8)}$          & 23.89$^{(10)}$         & 0.6751$^{(10)}$         & 0.4577$^{(8)}$*          \\
        \hline
        SRResNet~\cite{ledig_photo-realistic_2017} \textbf{[CVPR-17]}      & $\mathcal{L}_2$                                                                                                          & 25.82$^{(6)}$                      & 0.7369$^{(6)}$                    & -                       & 25.30$^{(6)}$          & 0.7297$^{(7)}$          & -                        \\
        \hline
        URDGN~\cite{yu_ultra-resolving_2016} \textbf{[ECCV-16]}            & $\mathcal{L}_2$ + $\mathcal{L}_{GAN}$                                                                                    & 24.63$^{(8)}$                      & 0.6851$^{(9)}$                    & -                       & 24.22$^{(9)}$          & 0.6909$^{(9)}$          & -                        \\
        \hline
        RDN~\cite{zhang_residual_2018} \textbf{[ECCV-18]}                  & $\mathcal{L}_1$                                                                                                          & 26.13$^{(5)}$                      & 0.7412$^{(5)}$                    & 0.1415$^{(4)}$          & 25.34$^{(5)}$          & 0.7249$^{(8)}$          & 0.4437$^{(7)}$*          \\
        \hline
        PFSR~\cite{kim_progressive_2019} \textbf{[BMVC-18]}                & $\mathcal{L}_2$ + $\mathcal{L}_{Perceptual}$ + $\mathcal{L}_{GAN}$ + $\mathcal{L}_{Heatmap}$ + $\mathcal{L}_{Attention}$ & 24.43$^{(9)}$                      & 0.6991$^{(8)}$                    & 0.1917$^{(7)}$          & 24.73$^{(8)}$          & 0.7323$^{(6)}$          & 0.3498$^{(4)}$*          \\
        \hline
        \multirow{2}{*}{FSRNet~\cite{chen_fsrnet_2018} \textbf{[CVPR-18]}} & $\mathcal{L}_2$ + $\mathcal{L}_{Perceptual}$                                                                             & 26.48$^{(3)}$                      & 0.7718$^{(3)}$                    & 0.1430$^{(5)}$          & 25.90$^{(4)}$          & 0.7759$^{(3)}$          & 0.3723$^{(6)}$*          \\
                                                                           & $\mathcal{L}_2$ + $\mathcal{L}_{Perceptual}$ + $\mathcal{L}_{GAN}$                                                       & 25.06$^{(7)}$                      & 0.7311$^{(7)}$                    & 0.1463$^{(6)}$          & 24.99$^{(7)}$          & 0.7424$^{(5)}$          & 0.3408$^{(3)}$*          \\
        \hline
        \multirow{2}{*}{DICNet~\cite{ma_deep_2020} \textbf{[CVPR-20]}}     & $\mathcal{L}_1$ + $\mathcal{L}_{Alignment}$                                                                              & 27.28$^{(2)}$                      & 0.7929$^{(2)}$                    & 0.1345$^{(3)}$          & 26.69$^{(2)}$          & 0.7933$^{(2)}$          & 0.3674$^{(5)}$*          \\
                                                                           & $\mathcal{L}_1$ + $\mathcal{L}_{Alignment}$ + $\mathcal{L}_{Perceptual}$ + $\mathcal{L}_{GAN}$                           & 26.34$^{(4)}$                      & 0.7562$^{(4)}$                    & 0.1319$^{(2)}$          & 25.96$^{(3)}$          & 0.7624$^{(4)}$          & \textbf{0.3336$^{(1)}$}* \\
        \hline
        Ours                                                 & w/o. Internal                                                       & \textbf{27.39$^{(1)}$}             & \textbf{0.7973$^{(1)}$}           & \textbf{0.1292$^{(1)}$} & \textbf{26.94$^{(1)}$} & \textbf{0.8005$^{(1)}$} & 0.3366$^{(2)}$*          \\
        \bottomrule
      \end{tabular}}
      \vspace{-10px}
  \end{table*}

\begin{table}[htbp]
    \centering
    \caption{Quantitative comparison on the natural image dehazing. Our proposed objective is capable of being extended to the dehazing task based on MSBDN-DFF, which shows superiority in both the indoor and outdoor datasets.}
    \resizebox{\linewidth}{!}{
      \begin{tabular}{lccccccccccc}
        \toprule
        Method                  & Metric         & DCP~\cite{he_single_2010} & MSCNN~\cite{ren_single_2016} & DcGAN~\cite{li_single_2018} & GFN~\cite{ren_gated_2018} & PFFNet~\cite{mei_progressive_2018} & GDN~\cite{liu_griddehazenet_2019} & DuRN~\cite{liu_dual_2019} & MSBDN-DFF~\cite{dong_multi-scale_2020} & Ours      \\
        \midrule
        \multirow{2}{*}{I-HAZE} & PSNR$\uparrow$ & 14.43$^{(9)}$            & 15.22$^{(8)}$                & 16.06$^{(5)}$               & 15.84$^{(7)}$             & 16.01$^{(6)}$                      & 16.62$^{(4)}$                     & 21.23$^{(3)}$             & 23.93$^{(2)}$                      & \textbf{24.31$^{(1)}$} \\
                                & SSIM$\uparrow$ & 0.752$^{(6)}$             & 0.755$^{(5)}$                & 0.733$^{(9)}$              & 0.751$^{(7)}$             & 0.740$^{(8)}$                      & 0.787$^{(4)}$                     & 0.842$^{(3)}$             & 0.891$^{(2)}$                      & \textbf{0.902$^{(1)}$} \\
        \hline
        \multirow{2}{*}{O-HAZE} & PSNR$\uparrow$ & 16.78$^{(9)}$            & 17.56$^{(8)}$                & 19.34$^{(4)}$               & 18.16$^{(7)}$             & 18.76$^{(6)}$                      & 18.92$^{(5)}$                     & 20.45$^{(3)}$             & 24.36$^{(2)}$                      & \textbf{24.79$^{(1)}$} \\
                                & SSIM$\uparrow$ & 0.653$^{(8)}$             & 0.650$^{(9)}$               & 0.681$^{(4)}$               & 0.671$^{(6)}$             & 0.669$^{(7)}$                      & 0.672$^{(5)}$                     & 0.688$^{(3)}$             & 0.749$^{(2)}$                      & \textbf{0.787$^{(1)}$} \\
        \bottomrule
      \end{tabular}}
      \label{tab:dehazing}
      \vspace{-10px}
  \end{table}

\subsection{Natural Image Restoration}
\begin{figure}[hbtp]
    \centering
    \begin{subfigure}[t]{.24\linewidth}
      \captionsetup{justification=centering, labelformat=empty, font=small}
      \subfloat[\#42 Hazy Image]{\includegraphics[width=0.49\linewidth]{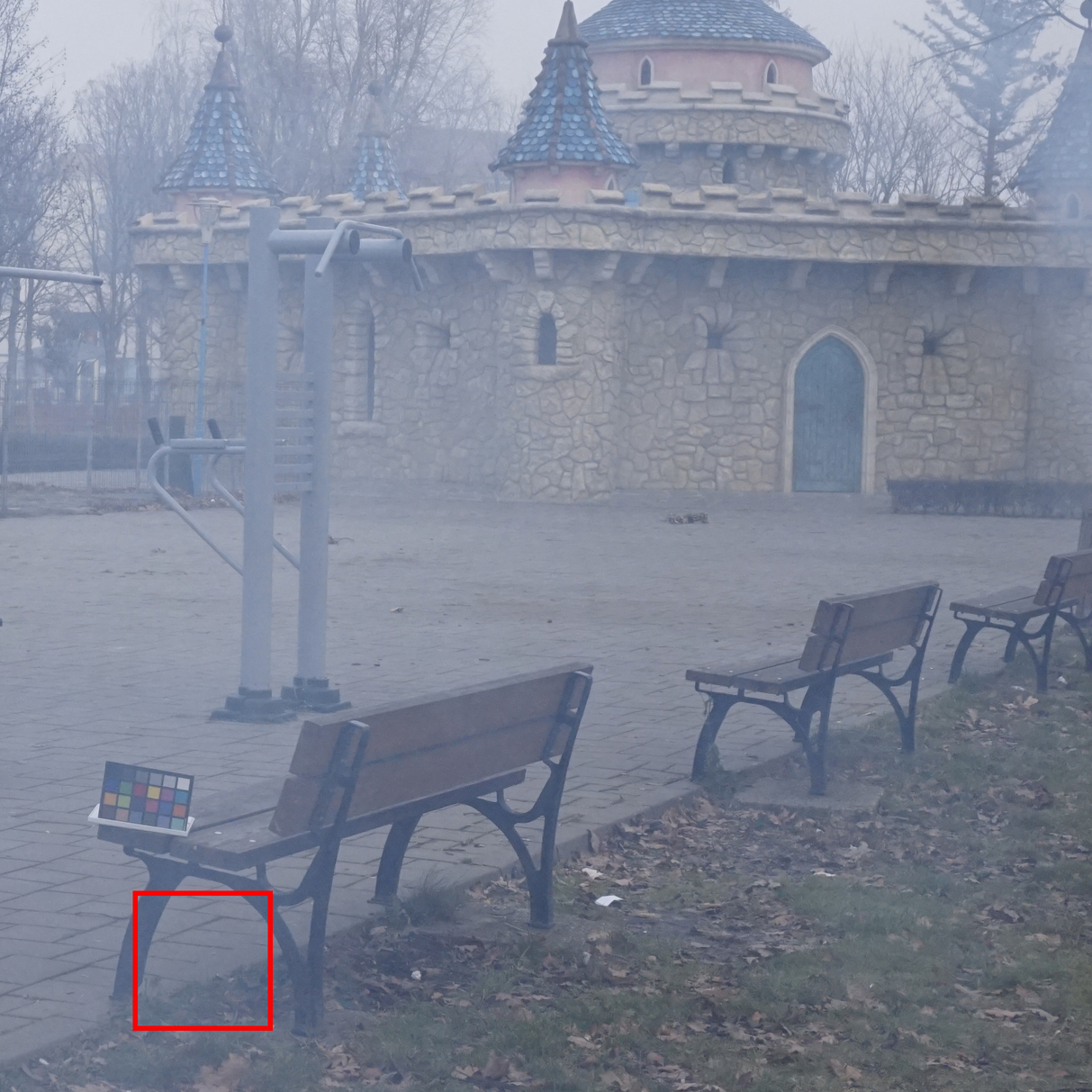}}\hfill
      \subfloat[MSBDN-DFF]{\includegraphics[width=0.49\linewidth]{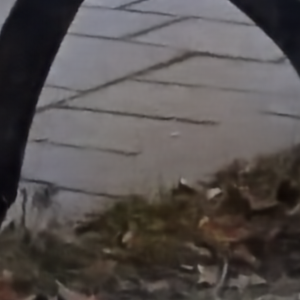}}\\
      \subfloat[Ours]{\includegraphics[width=0.49\linewidth]{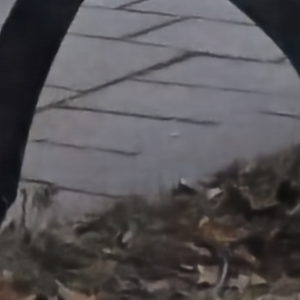}}\hfill
      \subfloat[GT]{\includegraphics[width=0.49\linewidth]{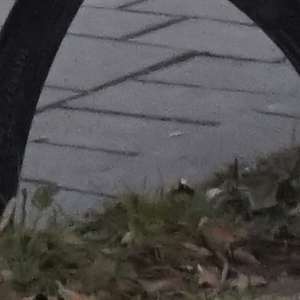}}
    \end{subfigure}%
    \hfill
    \begin{subfigure}[t]{.24\linewidth}
      \captionsetup{justification=centering, labelformat=empty, font=small}
      \subfloat[\#42 Hazy Image]{\includegraphics[width=0.49\linewidth]{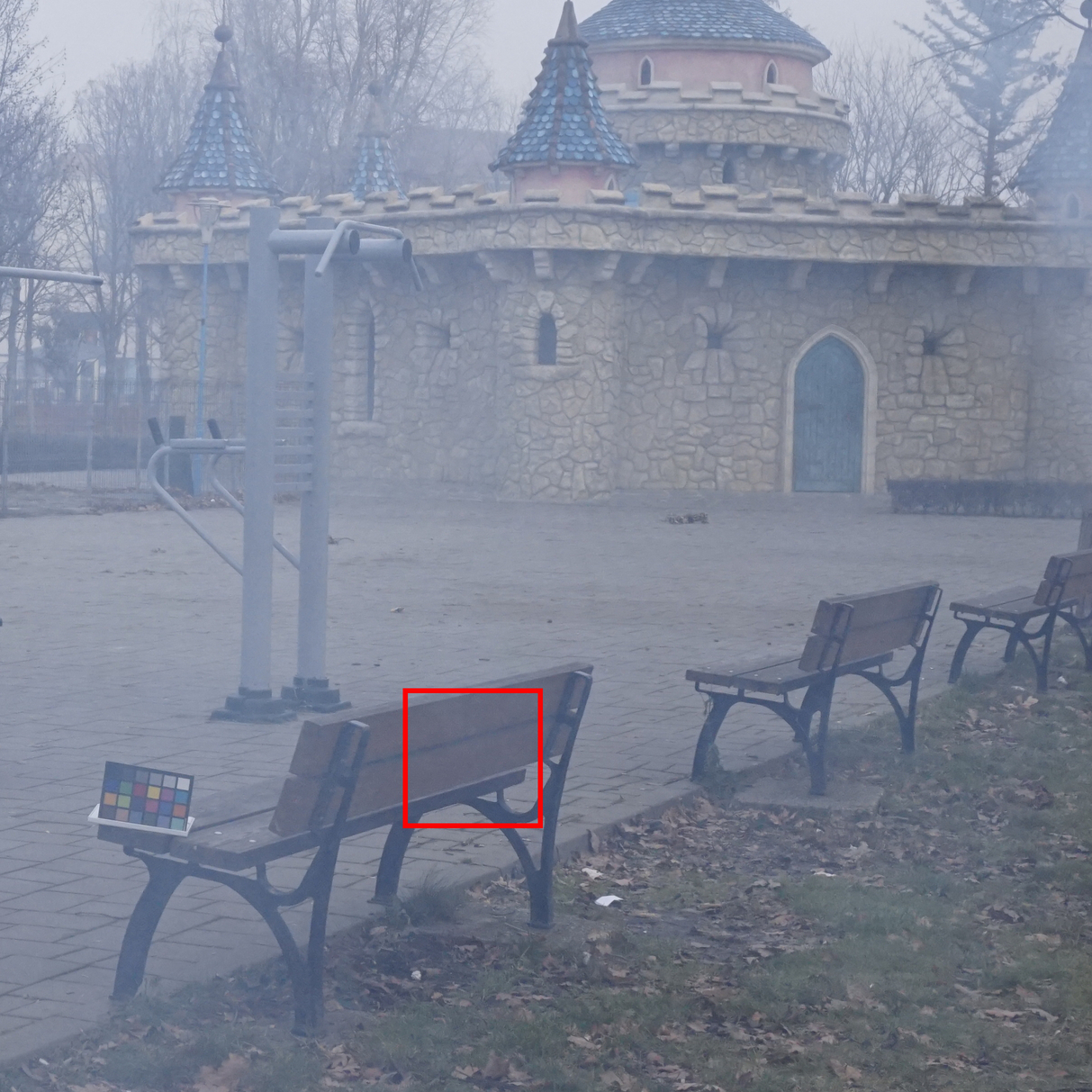}}\hfill
      \subfloat[MSBDN-DFF]{\includegraphics[width=0.49\linewidth]{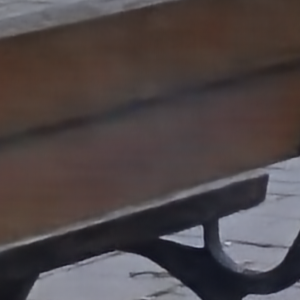}}\\
      \subfloat[Ours]{\includegraphics[width=0.49\linewidth]{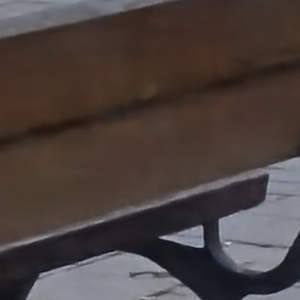}}\hfill
      \subfloat[GT]{\includegraphics[width=0.49\linewidth]{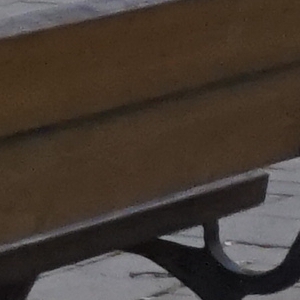}}
    \end{subfigure}%
    \hfill
    \begin{subfigure}[t]{.24\linewidth}
      \captionsetup{justification=centering, labelformat=empty, font=small}
      \subfloat[\#45 Hazy Image]{\includegraphics[width=0.49\linewidth]{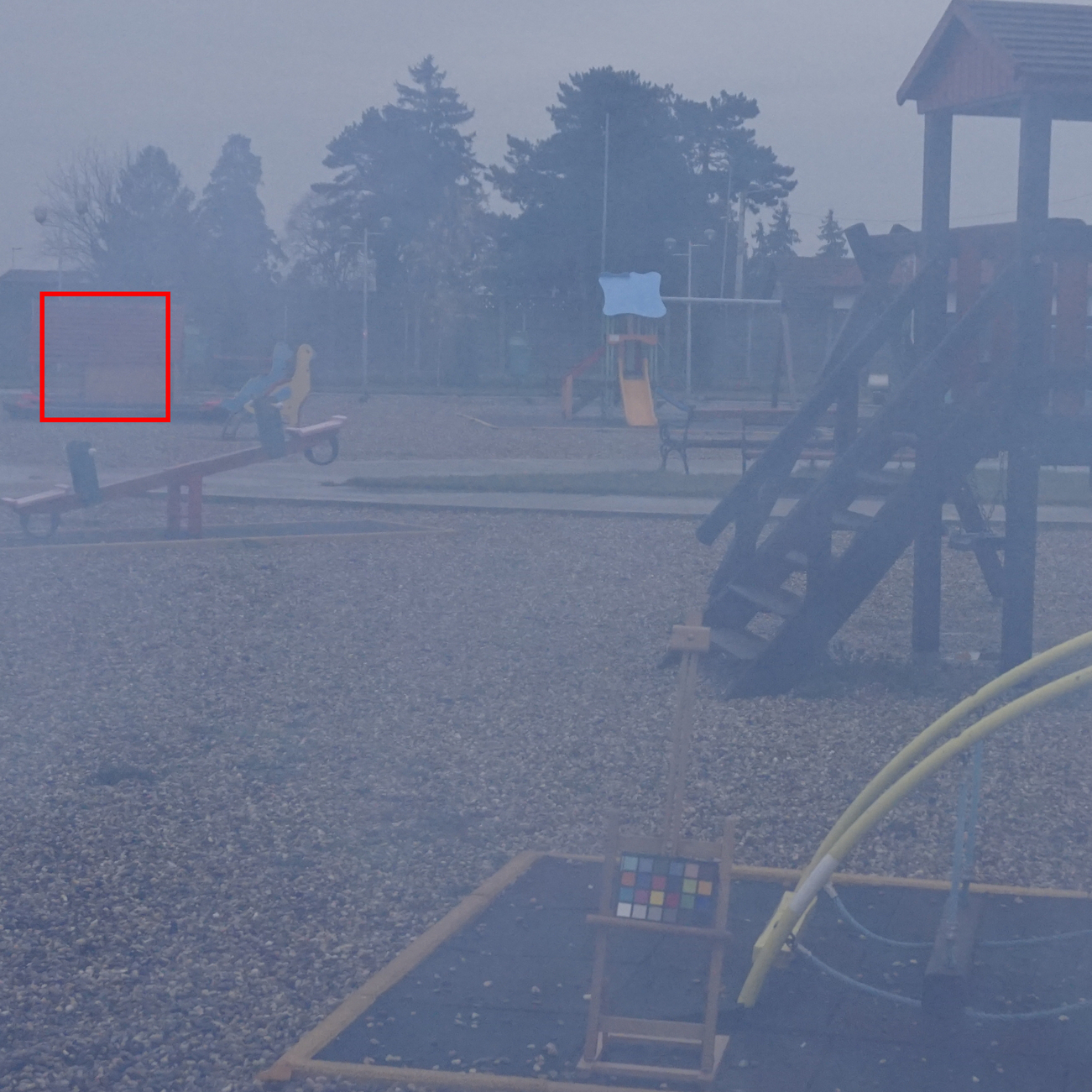}}\hfill
      \subfloat[MSBDN-DFF]{\includegraphics[width=0.49\linewidth]{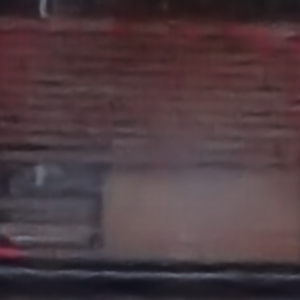}}\\
      \subfloat[Ours]{\includegraphics[width=0.49\linewidth]{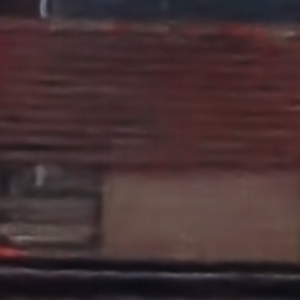}}\hfill
      \subfloat[GT]{\includegraphics[width=0.49\linewidth]{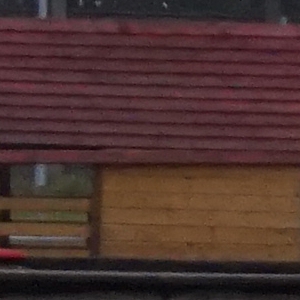}}
    \end{subfigure}%
    \hfill
    \begin{subfigure}[t]{.24\linewidth}
      \captionsetup{justification=centering, labelformat=empty, font=small}
      \subfloat[\#45 Hazy Image]{\includegraphics[width=0.49\linewidth]{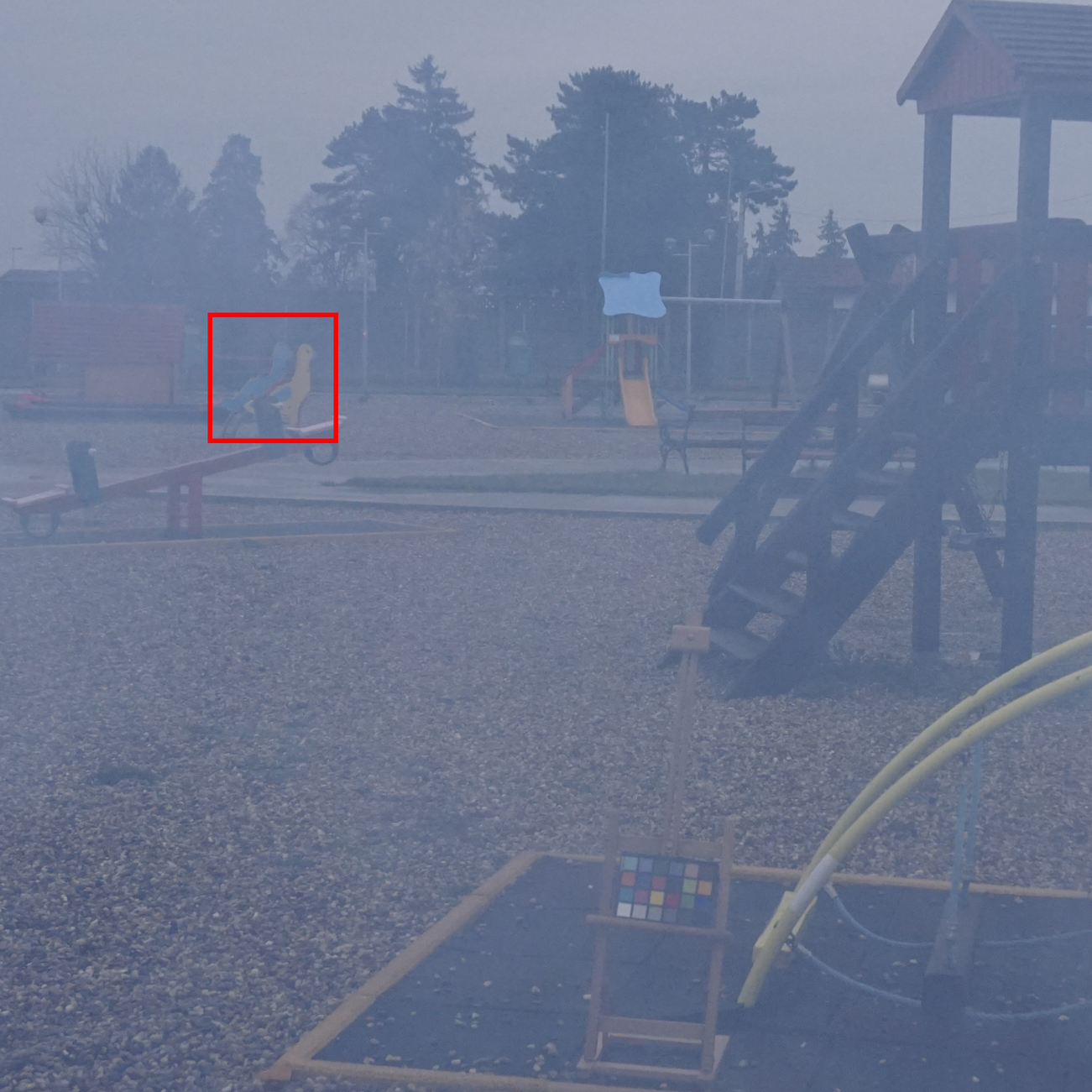}}\hfill
      \subfloat[MSBDN-DFF]{\includegraphics[width=0.49\linewidth]{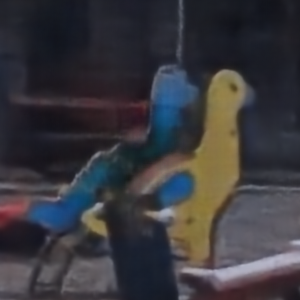}}\\
      \subfloat[Ours]{\includegraphics[width=0.49\linewidth]{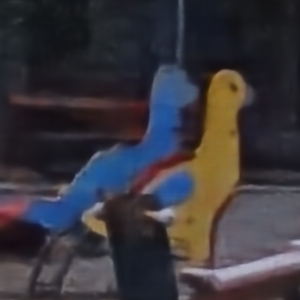}}\hfill
      \subfloat[GT]{\includegraphics[width=0.49\linewidth]{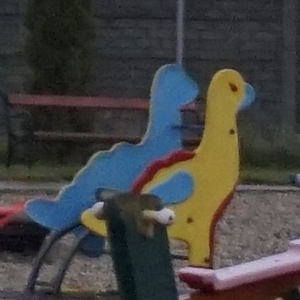}}
    \end{subfigure}%
    \caption{Qualitative comparison on the real-world dehazing. Compared with the SOTA method that employs pixel-wise loss functions, our extended version better recover the scenes under severe ill-posed distortion.}
    \label{fig:dehazing}
  \end{figure}

Different from cityscape images collected from limited scenes, natural images contain more diverse and complex semantics.
Therefore, the intrinsic semantics of natural images are more complex and diverse, which indicate a more challenging probability distribution estimation for our method.
To validate our effectiveness in such cases, here we follow the settings of the state-of-the-art dehazing method, \emph{i.e.}, MSBDN-DFF~\cite{dong_multi-scale_2020} on the end-to-end dehazing tasks~\cite{mei_progressive_2018}, and we extend the dehazing network with our proposed objective.
As the quantitative performance comparison shown in Table~\ref{tab:dehazing}, though the density estimation is challenging, our method can successfully outperform the compared method in both the indoor scenes and outdoor scenes without additional cost.
In Figure~\ref{fig:dehazing}, we show some randomly highlighted visual results for comparison, and all of our results contain the most clear appearance with less haze remained.
Specifically, we can notice that objects, \emph{e.g.}, \emph{floor}, \emph{chair}, \emph{roof}, \emph{toy} that contain certain semantics, are better restored with accurate color than the method trained with pixel-wise loss functions only.
This phenomenon further demonstrates the semantics transferring ability of our method, which regularizes restored objects to be semantic consistent and avoids incorrect color that againsts its semantics.

\begin{table}[htbp]
  \begin{varwidth}[b]{0.48\linewidth}
    \centering
    \resizebox{\linewidth}{!}{
    \begin{tabular}{lcccc}
      \toprule
      Method & $\sigma$ & PSNR $\uparrow$~ & SSIM $\uparrow$~ & MIoU (\%) $\uparrow$~ \\
      \midrule
      FFDNet & 25 & 35.03 & 0.925 & 0.605 \\
      + $\mathcal{L}_{iKLD}$ & 25 & 35.97 & 0.931 & 0.638 \\
      + $\mathcal{L}_{JSD}$ & 25 & 36.31 & 0.935 & 0.640  \\
      + $\mathcal{L}_{GAN}$ & 25 & 35.55 & 0.931 & 0.621  \\
      Ours & 25 & \textbf{36.45} & \textbf{0.936} & \textbf{0.644} \\
      \bottomrule
      \end{tabular}}
      \caption{Performance comparison with different distribution divergence.}
  \end{varwidth}%
  \hfill
  \begin{minipage}[b]{0.5\linewidth}
    \centering
    \includegraphics[width=.8\linewidth]{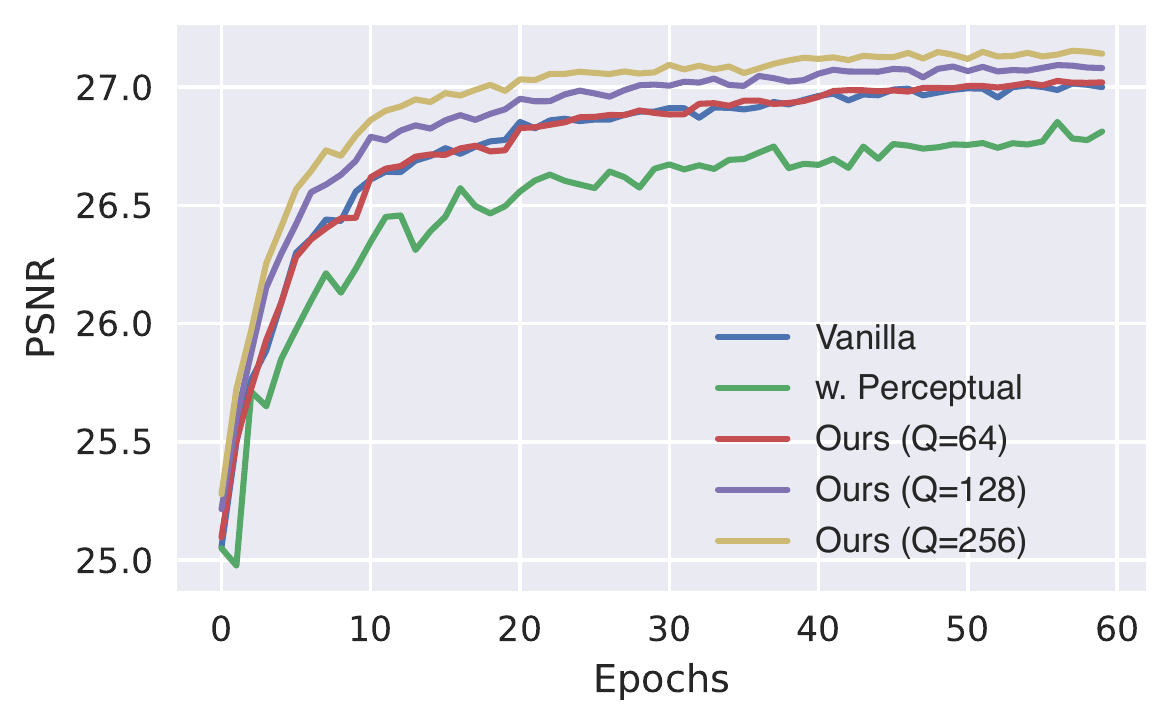}
    \captionof{figure}{Convergence visualization between different queue size.}
    \label{fig:convergence}
  \end{minipage}
\end{table}

\section{Discussion}
\paragraph{Designing Choice of KL Divergence.}
The way of the divergence estimation beween $\mathcal{G'}$ and $\mathcal{G}$ accounts a lot in our method.
We empirically employed the Kullback-Leibler (KL) divergence weighted by $\mathcal{G'}$ for D2SM.
Indeed, there are some other ways to estimate the divergence.
The most similar one, \emph{i.e.}, inverse KL divergence weighted by $\mathcal{G}$, which is also asymmetrical.
Based on KL divergence, another way to estimate the divergence is Jensen–Shannon divergence, which is symmetric and can be seen as the smoothed version of KL divergence, formulated as:
\begin{equation}
  D_{JS}(\mathcal{G'}||\mathcal{G}) = \frac{1}{2}D_{KL}(\mathcal{G'}||\mathcal{G}) + \frac{1}{2}D_{KL}(\mathcal{G}||\mathcal{G'}).
\end{equation}
According to \cite{arjovsky2017towards}, the optimization procedure of the optimal discriminator $D*$ in GAN yields minimizing the JS divergence, formulated as:
\begin{equation}
  \mathcal{L}(D*, G) = 2D_{JS}(\mathcal{G'}||\mathcal{G}) - 2\log2.
\end{equation}
Here we present the quantitative comparison with the three additional divergence estimation or objectives in Cityscapes.
The comprasion empirically proves our superiority of using KL-divergence compared with the others in the denoising tasks.

\paragraph{Designing Choice of Queue Size.}
The convergence curve visualized in the Figure~\ref{fig:convergence} further demonstrates that our proposed method significantly accelerates convergence with the proposed memorized historic sampling.
From the figure we can notice that the applied historic sampling with large queue size ($Q > 64$) can greately accelerate the learning, while the vanilla version can only achieve poor performance ($Q\le 64$).
Thus, in our practical implementation, we chose the queue size with the possible max value under the computational limitation.

\section{Conclusion}
We propose a simple but practical method for facilitating the restoration learning and preserving the semantic attribute.
It does not rely on any external information nor introduce additional parameters.
By implicitly approximating the divergence on the semantic feature space, we can force existed generation networks to learn to preserve semantic attributes during restoration learning.
We further transfer the method from the single-semantic image to the complex-semantic image \ie natural image by using internal statistics.
Empirically evaluation validates that the proposed method can be adapted to various restoration tasks and network architectures with general performance improvement.

\section*{Acknowledgments}
This work was supported in part by Shenzhen Science and Technology Program ZDSYS20211021111415025 and JCYJ20190813170601651.

\clearpage
%
%
\bibliographystyle{splncs04}
\bibliography{egbib}

\clearpage
\appendix
\section{Experimental Details}
Here we present the datasets used during our experiments, as well as their settings that may differ from the offical settings.

\begin{itemize}
  \item \textit{Cityscape Denoising and Segmentation}, which exploits Cityscapes~\cite{cordts_cityscapes_2016} dataset (2975 images for training and 500 images for testing) for denoising and segmentation. In regard to the semantic feature extraction, we use the penultimate layer of VGG19~\cite{simonyan_very_2015} pre-trained on the ImageNet~\cite{deng_imagenet_2009} dataset for object classification. For the high-level vision tasks, we use the HRNet48~\cite{wang_deep_2020} pre-trained on the Cityscapes dataset for semantic segmentation.
  
  \item \textit{Face Super-resolution and Alignment}, which exploits CelebA~\cite{liu_deep_2015} (168854 images for training and 1000 images for testing) and Helen~\cite{le_interactive_2012} (2005 images for training and 50 images for testing) datasets for face super-resolution and face landmark detection (denoted as alignment accuracy). In regard to the semantic feature extraction, we use the penultimate layer of LightCNN-9~\cite{wu_light_2018} pre-trained on the CelebA~\cite{liu_deep_2015} dataset for face recognition. For the high-level vision tasks, we use hourglass~\cite{newell_stacked_2016} networks pre-trained on the CelebA dataset for face landmark detection.

  \item \textit{Natural Image Restoration}, which exploits the real-world restoration datasets, \ie, i-haze~\cite{ancuti_i-haze_2018} (30 images for training and 5 images for testing) as well as o-haze~\cite{ancuti_o-haze_2018} (40 images for training and 5 images for testing) for dehazing. The setting of semantic feature extraction is same as the first experiment.
\end{itemize}

\end{document}